\documentclass{article}

\usepackage{arxiv}

\usepackage[utf8]{inputenc} 
\usepackage[T1]{fontenc}    
\usepackage{hyperref}       
\usepackage{url}            
\usepackage{booktabs}       
\usepackage{amsfonts}       
\usepackage{nicefrac}       
\usepackage{microtype}      
\usepackage{lipsum}		
\usepackage{graphicx}
\usepackage{doi}
\usepackage{bbm}
\usepackage{amssymb}
\usepackage{amsmath}
\usepackage{hyperref}

\usepackage{array}

\usepackage{algorithm}
\usepackage{float}
\usepackage{multirow}
\usepackage{caption}
\usepackage{subcaption}
\usepackage{algpseudocode}
\usepackage{enumitem}

\RequirePackage[normalem]{ulem} 
\RequirePackage{color}\definecolor{RED}{rgb}{1,0,0}\definecolor{BLUE}{rgb}{0,0,1} 

\usepackage{lineno}

\title{COMPASS: Computational Mapping of Patient-Therapist Alliance Strategies with Language Modeling}

\date{}

\author{Baihan Lin$^1$$^2$$^3$$^4$$^*$, Djallel Bouneffouf$^5$, Yulia Landa$^2$$^6$, Rachel Jespersen$^2$, Cheryl Corcoran$^2$$^6$, Guillermo Cecchi$^5$ \\
    $^1$ Department of Artificial Intelligence and Human Health, Icahn School of Medicine at Mount Sinai \\
	$^2$ Department of Psychiatry, Icahn School of Medicine at Mount Sinai \\
	$^3$ Department of Neuroscience, Icahn School of Medicine at Mount Sinai \\
	$^4$ Berkman Klein Center for Internet \& Society, Harvard University \\
	$^5$ IBM Research, T.J. Watson Research Center \\
    $^6$ Mental Illness Research, Education and Clinical Center, James J. Peters VA Medical Center \\
    $^*$ Corresponding Author: Baihan Lin, PhD ({baihan.lin@mssm.edu}) \\
}
\renewcommand{\headeright}{}
\renewcommand{\undertitle}{}
\renewcommand{\shorttitle}{Lin et al., \textit{Translational Psychiatry}, in press}

\hypersetup{
pdftitle={COMPASS: Computational Mapping of Patient-Therapist Alliance Strategies with Language Modeling},
pdfsubject={q-bio.NC, q-bio.QM},
pdfauthor={Baihan Lin},
pdfkeywords={First keyword, Second keyword, More},
}

\begin{document}
\maketitle
\begin{abstract}

The therapeutic working alliance is a critical predictor of psychotherapy success. Traditionally, working alliance assessment relies on questionnaires completed by both therapists and patients. In this paper, we present COMPASS, a novel framework to directly infer the therapeutic working alliance from the natural language used in psychotherapy sessions. Our approach leverages advanced large language models (LLMs) to analyze session transcripts and map them to distributed representations. These representations capture the semantic similarities between the dialogues and psychometric instruments, such as the Working Alliance Inventory. Analyzing a dataset of over 950 sessions spanning diverse psychiatric conditions -- including anxiety (N=498), depression (N=377), schizophrenia (N=71), and suicidal tendencies (N=12) -- collected between 1970 and 2012, we demonstrate the effectiveness of our method in providing fine-grained mapping of patient-therapist alignment trajectories, offering interpretable insights for clinical practice, and identifying emerging patterns related to the condition being treated. By employing various deep learning-based topic modeling techniques in combination with prompting generative language models, we analyze the topical characteristics of different psychiatric conditions and how these topics evolve during each turn of the conversation. This integrated framework enhances the understanding of therapeutic interactions, enables timely feedback for therapists on the quality of therapeutic relationships, and provides clear, actionable insights to improve the effectiveness of psychotherapy.
\end{abstract}


\section{Introduction}\label{sec1}

The working alliance, which encompasses various cognitive and emotional aspects of the therapist-patient relationship, is a critical concept in psychotherapy identified as a crucial factor in predicting treatment outcomes \cite{Bordin79,Wampold2015}. Research over the past few decades has consistently demonstrated the significant role that a strong working alliance plays in determining the success of various therapeutic modalities, especially in patient engagement and adherence to treatment \cite{horvath1994working,norcross2010therapeutic,eliacin2018relationship}. However, current methods for assessing the alliance rely on evaluating entire therapy sessions using point-scale ratings \cite{horvath1981exploratory}, with questionnaires and clinical instruments such as the working alliance inventory (WAI). As a current standard in the field, the working alliance inventory (WAI) is a self-report measurement designed to quantify the therapeutic bond, task agreement, and goal agreement in psychotherapy \cite{horvath1981exploratory,tracey1989factor,martin2000relation}. It is typically completed by both patient and therapist at multiple points during therapy to monitor the development and quality of the therapeutic alliance. Common practice involves administering the WAI at the beginning of treatment to establish a baseline, at mid-treatment to track progress, and toward the end to assess changes in the alliance over time. In some settings, it may also be completed periodically throughout therapy, such as after every few sessions or even after each session, to continuously evaluate alliance strength.

The WAI has been widely used to assess the quality of the working alliance between therapists and patients, demonstrating strong psychometric properties \cite{tracey1989factor,martin2000relation}. Research indicates that higher WAI scores are associated with better therapy outcomes across various diagnoses, including depression and anxiety disorders. The effect sizes for the correlation between alliance quality and treatment outcomes range from 0.19 to 0.32, suggesting a moderate relationship \cite{ardito2011therapeutic,paap2022working}. The WAI’s reliability is evidenced by a high internal consistency (Cronbach's alpha around 0.93) and validation across diverse populations and therapeutic settings \cite{ardito2011therapeutic,hunik2021exploring}.

Alliance development patterns can vary based on the therapeutic approach. For instance, in cognitive-behavioral therapy (CBT), a strong initial alliance is often crucial for patient engagement and treatment adherence \cite{hunik2021exploring}. Studies have also noted differences in alliance strength and its associations with treatment outcomes across diagnostic groups. A study comparing alliance ratings among patients with depression, somatoform disorders, and eating disorders found no significant differences in the strength of alliance ratings or their associations with treatment outcomes across these groups. All three groups reported positive alliances that improved over the course of therapy \cite{mander2017therapeutic}. In all studied disorder groups, there was a notable incongruence between patient and therapist ratings of the therapeutic alliance. This discrepancy highlights the importance of aligning perceptions to enhance therapeutic processes and outcomes. Clients with severe mental illnesses, such as schizophrenia or personality disorders, often rate the alliance higher than their therapists do \cite{igra2020meta}. This discrepancy can impact therapy dynamics and outcomes \cite{igra2020meta}. For instance, clients with substance misuse issues tend to exhibit larger rating discrepancies compared to those with other severe disturbances. Patterns of alliance development can differ based on diagnosis, with some studies identifying stable alliances or linear growth patterns as being correlated with better outcomes \cite{ardito2011therapeutic}. Understanding these patterns can help tailor therapeutic approaches to individual needs.

For evidence-based psychotherapies such as cognitive behavioral therapy (CBT) \cite{dobson2018evidence}, quantifying best practices on a large scale has proven challenging. While CBT has demonstrated efficacy for various mental health conditions, including anxiety, depression, and psychosis, capturing the nuances of effective therapeutic techniques in real-world settings remains complex \cite{mcevoy2014relationship}. Traditional assessment methods fall short in providing the necessary granularity and scalability to capture the subtleties of therapist-patient interactions within sessions, which can be important, particularly, during child and adolescent developmental stages \cite{fernandes2022therapeutic}. The therapeutic alliance in CBT not only enhances engagement but also facilitates collaborative empiricism, a key principle where both therapist and client actively explore thoughts and behaviors together \cite{dobson2023therapeutic}. By examining alliance patterns within sessions, therapists can make real-time adjustments to interventions, refining techniques to meet individual needs. Hence, understanding this process via alliance can allow us to better quantify the best practice to generalize across populations. This is similarly true in other therapy settings, such as psychodynamic therapy and psychotherapy in general. For instance, in psychodynamic therapy, which often delves into unconscious processes and past experiences, the therapeutic alliance is particularly vital. The quality of this alliance can significantly affect the patient's willingness to engage with challenging material related to their emotions and relationships. A positive alliance can facilitate the exploration of transference, where feelings about significant others are projected onto the therapist—enabling patients to gain insights into their relational patterns \cite{ardito2011therapeutic}.

\begin{figure}[tb]
\centering
    \includegraphics[width=\linewidth]{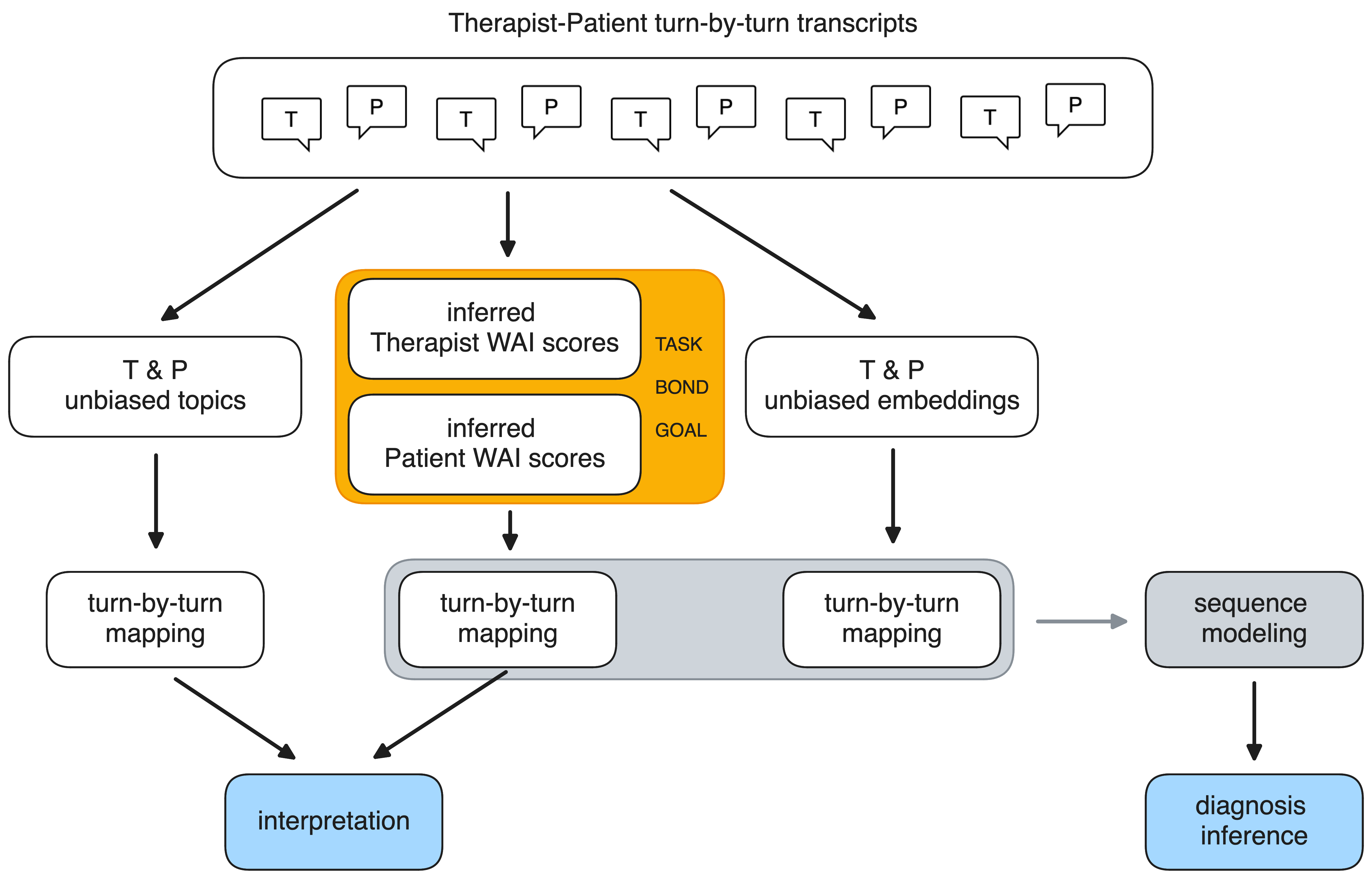}
\caption{\textbf{Analytical pipeline of the working alliance analysis.} The transcript is separated into turns by the therapist (T) and turns by the patients (P). These dyads of turns are compared separately by the working alliance inventories (WAI) for the clients and the therapists in the sentence embedding space, and the inferred WAI scores according to different inventory items are computed and then summarized into separate scales for Task, Bond and Goal. Topics and embeddings not biased by WAI are also computed for further analysis and interpretation through sequence modeling, a validating example of which is the diagnosis of psychiatric condition being treated only given the linguistic features of patient-doctor conversations.
}\label{fig:pipeline}
\end{figure}

While the WAI is a valuable tool, its periodic administration cannot capture the nuanced, moment-to-moment alliance dynamics that drive therapeutic progress. Furthermore, regular WAI assessment proves too time-intensive for both clinical practice and large-scale research. Although psychotherapy was an early adopter of Natural Language Processing (NLP), with pioneering efforts like ELIZA and Parry simulating therapeutic interactions \cite{shum2018eliza,zemvcik2019brief}, these innovations never gained systematic clinical adoption.

We propose leveraging large language models (LLMs) to quantify patient-therapist alliance by mapping dialogue turns onto established WAI dimensions \cite{horvath1981exploratory,tracey1989factor,martin2000relation}. Building on recent NLP advances \cite{rezaii2022natural,lin2022deep,lin2022working,lin2022neural,valliance}, our approach enables analysis of alliance patterns across multiple timescales - from individual turns to entire therapeutic trajectories.
Our methodology, COMPASS (\underline{\textbf{CO}}mputational \underline{\textbf{M}}apping of \underline{\textbf{P}}atient-Therapist \underline{\textbf{A}}lliance \underline{\textbf{S}}trategie\underline{\textbf{S}}), analyzes therapy transcripts by mapping each conversational turn to WAI concepts, generating granular alliance scores that can reveal diagnosis-specific patterns in therapeutic bond development \cite{lin2022working}. We validate this approach using the Alexander Street dataset \cite{alexstreet}, comprising over 950 transcribed sessions with patients diagnosed with anxiety (N=498), depression (N=377), schizophrenia (N=71), and suicidal tendencies (N=12).
Previous research has demonstrated NLP's utility in analyzing mental health discourse, from uncovering latent structures in depression-related language on Twitter \cite{resnik2015beyond} to enhancing PTSD detection \cite{zeng2012synonym}. 

Building on this foundation, we conduct systematic topic modeling at the dialogue turn level to generate interpretable insights (Figure \ref{fig:pipeline}). COMPASS's effectiveness is validated through improved classification and diagnostic capabilities compared to baseline models, offering actionable insights for enhancing psychotherapy strategies through timely, granular alliance monitoring.

\section{Methods and Materials}

\subsection{Psychotherapy Transcript Dataset}

We begin by introducing the dataset used in our study. The Alexander Street \textit{Counseling and Psychotherapy Transcripts} dataset \cite{alexstreet} consists of transcribed recordings of over 950 therapy sessions between multiple anonymized therapists and patients. Each session covers one of four psychiatric conditions: anxiety, depression, schizophrenia, and suicidal tendencies. This comprehensive collection includes speech-translated transcripts of the recordings from real therapy sessions, 40,000 pages of client narratives, and 25,000 pages of reference works.

The Alexander Street data includes a large repository of anonymized transcripts of therapy sessions, representing a wide range of approaches including prominently Group Counseling, Cognitive Behavior Therapy, Person-centered Therapy, Brief Relational Therapy, Relational Emotive Behavior Therapy, Psychoanalysis, and Integrative Psychoanalysis, spanning sessions conducted from 1970 through 2012, with client ranging in age from 11 to 80+ yo, an approximate female-male distribution of 70\%-30\% and less than 10\% bisexual/non-binary gender; the majority of therapists/counselors are anonymous (75\%+).

This dataset provides a comprehensive source for analyzing the therapeutic process in psychotherapy, where we extracted the following cohort with 498 sessions for anxiety, 377 for depression, 71 for schizophrenia, and 12 for suicidal tendencies. The sessions cover four types of psychiatric conditions: anxiety, depression, schizophrenia, and suicidal. Each dialogue pair consists of a patient response turn $S^p_i$ followed by a therapist response turn $S^t_i$. In total, the dataset contains over 200,000 turns from both patients and therapists, providing a rich source for analyzing the therapeutic process in psychotherapy.

\subsection{Deep Learning Inference of Working Alliance}

The analytic framework for inferring the working alliance from psychotherapy sessions is illustrated in Figure \ref{fig:pipeline}. Our approach involves analyzing the complete set of transcript records, which can be segmented based on timestamps or topic turns, from individual patients with various clinical conditions or from cohorts of patients with the same condition. The original data is presented in pairs of dialogues, and we extract features in three different ways: (1) using the full pairs of dialogues, (2) extracting only the patients' responses, or (3) extracting only the therapists' responses. Each feature set has its advantages and disadvantages. The dialogue features contain all the information but can mix together the intents within sentences from both individuals. 


The dialogue between the patient and therapist in a session is transcribed into pairs corresponding to the patient's turn, followed by the therapist's turn \footnote{This is of course under the assumption that, therapists often provide responses that are broad and generalized, affirming or summarizing the patient's input. These responses can be thought of as semantic ``labels'' that can be anticipated based on the ``inputs'' of the patient's statements. In reality, the patient-doctor dialogue could as well be initiated by the doctor. For the analytical purposes, we set the default to be patient first, without loss of generalizability to a lag of at most one turn.}. The inventories of working alliance questionnaires are also provided in pairs for the patient and the therapist, each comprising 36 statements. We employ sentence or paragraph embeddings to encode both the dialogue turns and the inventories; the embeddings are vectorial representations of text \cite{deerwester1990} that we then use to compute the similarity between turns and inventory item. This approach yields a 36-dimensional {\sl inferred} working alliance score for each patient and therapist turn; we  will further discuss the specific scales of our inferred working alliance scores in Section \ref{sec:wai} (see Supplementary Materials).

\subsection{Sentence Embeddings}
To represent the dialogue turns and working alliance inventories, we employ deep sentence embeddings. In this study, we use two types of sentence embeddings, Doc2Vec and SentenceBERT, which are two popular choices of deep learning-based embedding models of documents or sentences. We used these two models of different neural architectures to demonstrate the model-agnostic feasibility of the fine-grained linguistic analytics.

Doc2Vec \cite{le2014distributed} is an unsupervised learning model that learns vector representations of sentences and documents. It extends the traditional bag-of-words representation by incorporating a distributed memory that captures the context of the sentence. We use Doc2Vec to generate embeddings of the dialogue turns and working alliance inventories, resulting in 300-dimensional vectors.

SentenceBERT \cite{reimers2019sentence} is a modified version of the BERT model \cite{devlin2018bert} specifically designed for sentence embeddings. It utilizes siamese and triplet network structures to infer semantically meaningful sentence representations. We use SentenceBERT to generate 384-dimensional embeddings of the dialogue turns and inventories, which we use to obtain a 36-dimensional working alliance score for each turn, as described above, but also as agnostic representations of the turns  {\sl unbiased} by the working alliance inventory.

\subsection{Working Alliance Inventory}
\label{sec:wai}

The modern version of the working alliance inventory (WAI) consists of 36 questions, and participants are asked to rate each item on a 7-point scale (1=never, 7=always) \cite{martin2000relation}, as detailed by the full inventory in the supplementary material (Table \ref{tab:wai}). The working alliance inventories are usually administered with two versions, a client / patient version, and a therapist version. The inventory items are very similar with only a minor change of phrasing. The inventory aims to measure alliance factors across different types of therapy, establish the relationship between the alliance measure and theoretical constructs underlying the measure, and relate the alliance measure to a unified theory of therapeutic change \cite{horvath1994working}. Both patients and therapists fill out the WAI independently after each session, and responses are blinded to the other party to ensure unbiased assessments.

The 36 items of the WAI are used to derive three alliance scales: Task (the agreement on therapy-related tasks), Bond (the affective bond between therapist and patient), and Goal (the agreement on treatment goals). These scales capture the collaborative nature of the patient-therapist relationship, the affective bond between therapist and patient, and the agreement on treatment-related tasks and long-term goals \cite{horvath1994working}. Example items include: ``The therapist and I agree on what is important for me to work on'' (Task), ``I feel that the therapist likes me'' (Bond), and ``We agree on the steps to be taken to improve my situation'' (Goal). Each scale score is computed using a weighting matrix that assigns weights to the questionnaire responses to these inventory items based on a key table, resulting in a comprehensive assessment of the working alliance. Additionally, the overall working alliance score is obtained by summing the scores of the three scales \cite{horvath1994working}.

The assessment of WAI is largely restricted to research studies and has not been adopted in the regular practice of therapy, given it is relatively burdensome for both patients and therapist. Even in studies, it is not feasible for the participants to provide analysis of alliance at the turn level, something usually not considered or assessed by third parties. In contrast, we implemented such turn-by-turn analysis, relying on the proven capabilities of large language models to identify nuanced semantic content \cite{chang2024}, including the possibility of mapping self-evaluated inventories onto semi-structured interviews and monologues \cite{srivastava2023}. 

To illustrate how the process works, here's an example of three possible utterances occurring in a psychotherapy session (Table \ref{tab:example_wainference}). 
We can see the first sentence is clearly discussing the BOND aspect of the therapeutic relationship, where the client feels understood and supported by the therapist. The second sentence highlights the TASK aspect, focusing on the daily tasks that will help the client manage their anxiety as a GOAL. The third sentence emphasizes the GOAL aspect, reflecting the client's commitment to working towards better communication with their partner.

To quantify these aspects, we first compute their sentence embeddings in a high-dimensional space (300 dimensions in this case, using a Doc2Vec model). For instance, the first sentence's embedding is computed to be [-0.11, 0.05, 0.08,...], the second sentence become [0.05, 0.02, 0.12,...], and the third [-0.01, 0.10, -0.08,...]. Additionally, we have the working alliance inventories (36 items/questions) in their respective 300-dimension embedding space. For example, the first item "I felt uncomfortable with \_." corresponds to [0.01, 0.02, 0.01,...]. 

We then compute the sentence similarity (e.g., cosine similarity) of the 300-dimension embedding vector of the example sentence with the 300-dimension embedding vector of the inventory item, yielding a 36-dimensional similarity score. For example, the first sentence corresponds to [0.52, 0.49, 0.47, ...], with 0.52 being the similarity between the first sentence and the first inventory item. Using the key table corresponding to the working alliance scores, we map the 36 items to the 3 scales by calculating the dot product between the 36-dimensional similarity vector and the corresponding key table vector (e.g., [-1, 1, 0, 1, +1, 0, ...]). That is, the semantic similarity of the turns with respect to each item is summed up following the original aggregation of the WAI score (Table \ref{tab:wai}). This mapping produces the inferred scores for Task, Bond, and Goal, as presented in Table \ref{tab:example_wainference}.

\begin{table}[tb]
      \caption{\textbf{Examples of working alliance inference.} The semantic similarities of the sentences to the scales' items are summed up following the original WAI score aggregation (\ref{tab:wai}).
      }
      \label{tab:example_wainference} 
      \centering
 \begin{tabular}{ p{0.55\linewidth} | p{0.1\linewidth} | p{0.1\linewidth} | p{0.1\linewidth} }
Example sentences & Inferred BOND & Inferred TASK & Inferred GOAL \\ \hline
1. ``I feel really understood and supported by you. It's comforting to know I can talk about anything here without being judged.'' &  1.41 & -0.76 & -0.66 \\ \hline
2. ``I understand that practicing these relaxation techniques as daily tasks will help me manage my anxiety better.'' & -1.39 & 0.46 & 0.93 \\ \hline
3. ``I'm committed to working towards better communication with my partner, as we discussed in our last session.''  &  -0.85 & -0.56 & 1.40 \\ \hline
\end{tabular}
\end{table}

As expected, the first sentence shows a high score for BOND, the second for TASK and GOAL, and the third for GOAL, demonstrating the model's ability to differentiate and quantify the various aspects of the therapeutic working alliance.

 \begin{figure}[tb]
\centering
\includegraphics[width=\linewidth]{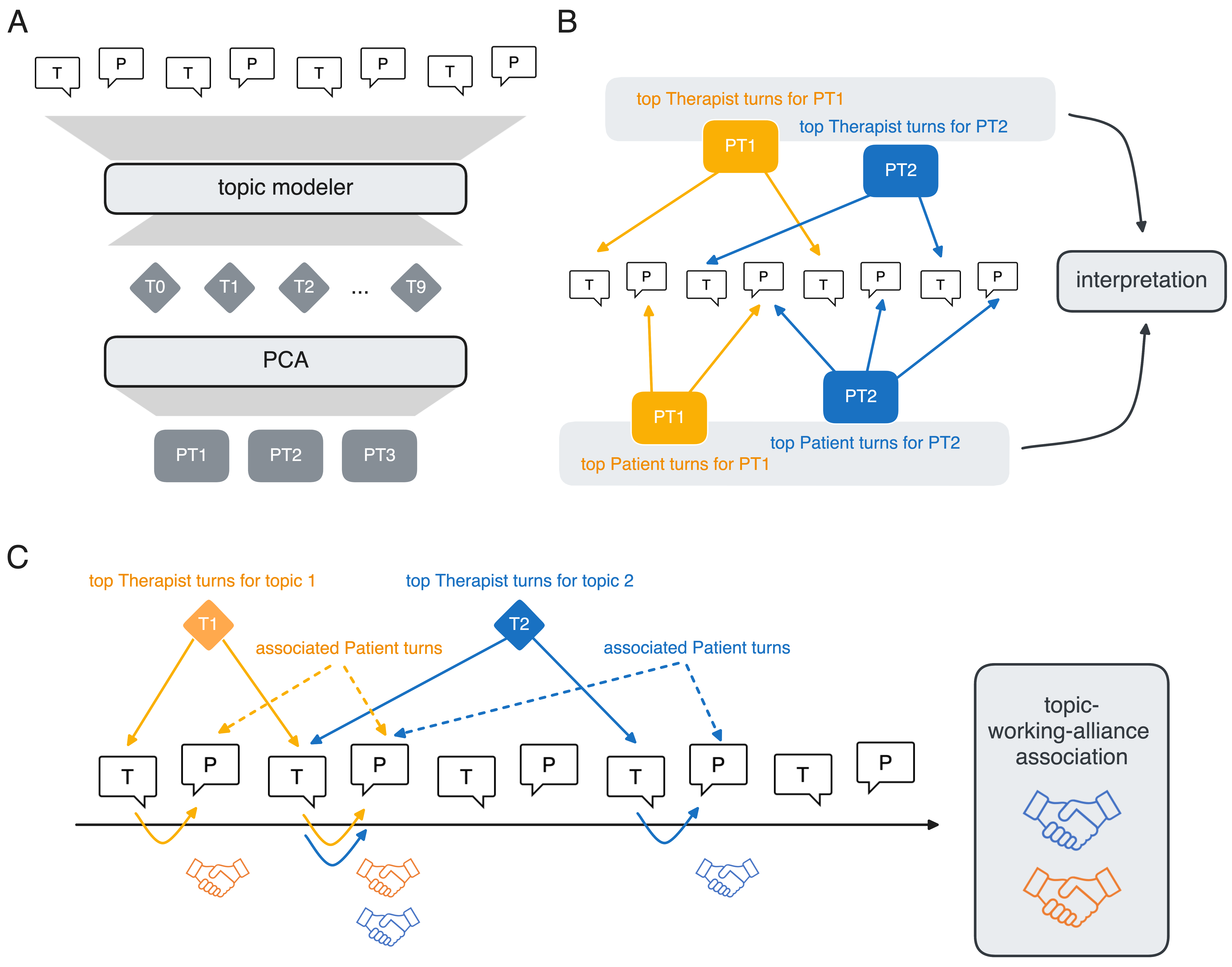}
\caption{\textbf{Flowcharts for topic modeling pipelines.} (A)  Topics are extracted from the sessions including therapists and patients turns. To facilitate additional interpretability, we coarse-grained the topics using PCA. (B) {Flowchart for identifying principal topics and their interpretation.} The turns with largest projections on the principal topics are fed into the language modeling interpreter to gain insights. (C) {Flowchart for identifying association between therapist topics and inferred patient working alliance.} The estimation of how inferred patient working alliance is conditioned by the therapist: top therapist turns for each topic are used to select the corresponding patient turns.}\label{fig:topic_flowchart}
\end{figure}

\subsection{Identification of Working Alliance Interaction Patterns}

In addition to the analytical features enabled by the working alliance analysis, we explore the usefulness of these features to identify patterns of interactions between patients and therapists, which we hypothesize emerge distinctively in the different conditions being treated. To this end, we implemented a Transformer-based neural architecture \cite{vaswani2017attention}, and a Long Short-Term Memory Network (LSTM) model \cite{hochreiter1997long}.

Specifically, we concatenate the 36-dimensional working alliance scores estimated from the current turn, as described above, with the unbiased sentence embedding of the turn. This combined feature vector is then fed into a sequence classifier, which we term Working Alliance Transformer (WAT) and Working Alliance LSTM (WA-LSTM) (see Supplementary Materials). By applying the WAT or the WA-LSTM to the psychotherapy transcripts, we can classify the clinical condition of the sequence based on the working alliance scores and the content of the dialogue turns. This classification model can be applied to the entirety of a session or a segment of the session; its accuracy provides a validation that these patterns can be identified, as we show in the Results section.

\subsection{Psychotherapy Topic Modeling Framework}

Topic modeling is a statistical technique used to uncover the latent semantic structures in a collection of documents. In the context of psychotherapy transcripts, topic modeling can reveal the underlying themes and topics discussed during therapy sessions, as well as provide additional insights in correlation with the therapeutic alliance between the patient and therapist.

While classical topic modeling approaches have shown effectiveness in the past, recent advancements in deep learning have led to the emergence of Neural Topic Modeling as a superior solution compared to its classical counterparts in terms of its representational power \cite{miao2016neural}. In this context, we propose the utilization of Neural Topic Modeling \cite{lin2022neural} to uncover the topical propensities associated with different psychiatric conditions using psychotherapy session transcripts. Furthermore, we incorporate temporal modeling techniques to provide additional interpretability. 

The full topic modeling pipeline is illustrated in Figure \ref{fig:topic_flowchart}A. By applying these neural topic models to the psychotherapy transcripts, we can uncover the latent topics discussed during therapy sessions. These topics provide valuable insights into the content and focus of the therapy, allowing for a more comprehensive understanding of the therapeutic process. The goal is to uncover the top 10 topics and extract more distinctive features for subsequent tasks. To accomplish this, a principal component analysis (PCA) is conducted on the topic space to extract a coarse-grained representation. Through this analysis, three principal topic spaces are identified, which encompass the patient turns and the corresponding therapeutic interventions undertaken by the therapists. 

As illustrated in Figure \ref{fig:topic_flowchart}B, to interpret these topics, we select the top turns of the therapist and patient dialogue ranked by the topic scores, and use a generative Large Language Model (LLM), ChatGPT based on GPT-3.5, to provide an interpretation by prompting it for summaries of the topics given the top turns that most exemplify the topic, as follows:
``I have the following top sentences exemplifying three principal topic spaces. Can you summarize what the three topics the patients are talking about, respectively?'', and ``Again, I have the following top sentences exemplifying the three principal topic spaces. Can you summarize what the three intervention items attributed to each principal topic spaces the therapists are talking about, respectively? For instance, what therapeutic interventions is the therapist applying.''
This allows us to expand interpretability possibilities, and diminish the effect of our biases as researchers. 

Using a language model to interpret text data offers several advantages for researchers and practitioners: (1) the model can provide an consistent analysis of the data that, while necessarily incomplete, is devoid of individual biases that analysts might inadvertently introduce; (2) by relying on the model's interpretation, researchers can access a more neutral perspective, enhancing the objectivity of their findings; (3) language models can quickly process and analyze large volumes of text, identifying patterns, relationships, and insights that may be challenging for humans to detect manually, and eventually guide and supplement the daily practice of therapy.

\begin{figure}[tb]
\centering
    \includegraphics[width=\linewidth]{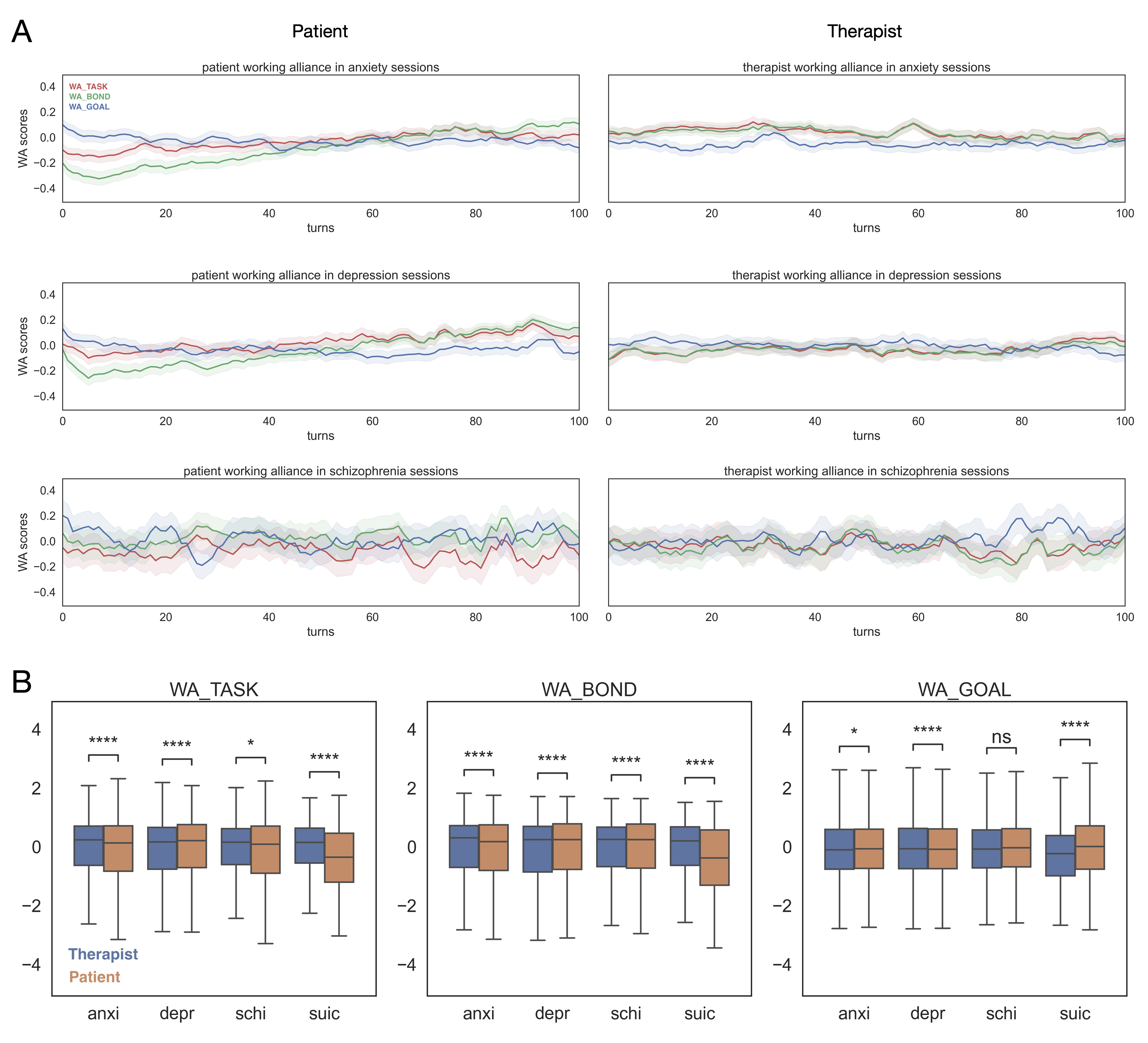} 
\caption{\textbf{Working alliance scores in the patient and therapist sessions of different clinical conditions.} After standardizing the working alliance scores, we pooled the sessions into different psychiatric conditions and averaged the working alliance scores of the patients and therapists separately at each time step (i.e. dialogue turn). (A) The progression of the working alliance over the sessions can be observed as well as their distinctions across the clinical conditions their corresponding session belong to. (B) The differences between the working alliance scores of therapist and patient turns are also highlighted in boxplots, tested with T-test for the means of the two independent samples of scores (p-value notations: **** 1e-4, *** 1e-3, ** 1e-2, * 0.05, ns for ``not significant'').
}\label{fig:avg_score}
\end{figure}

 \begin{figure}[tb]
\centering
    \includegraphics[width=\linewidth]{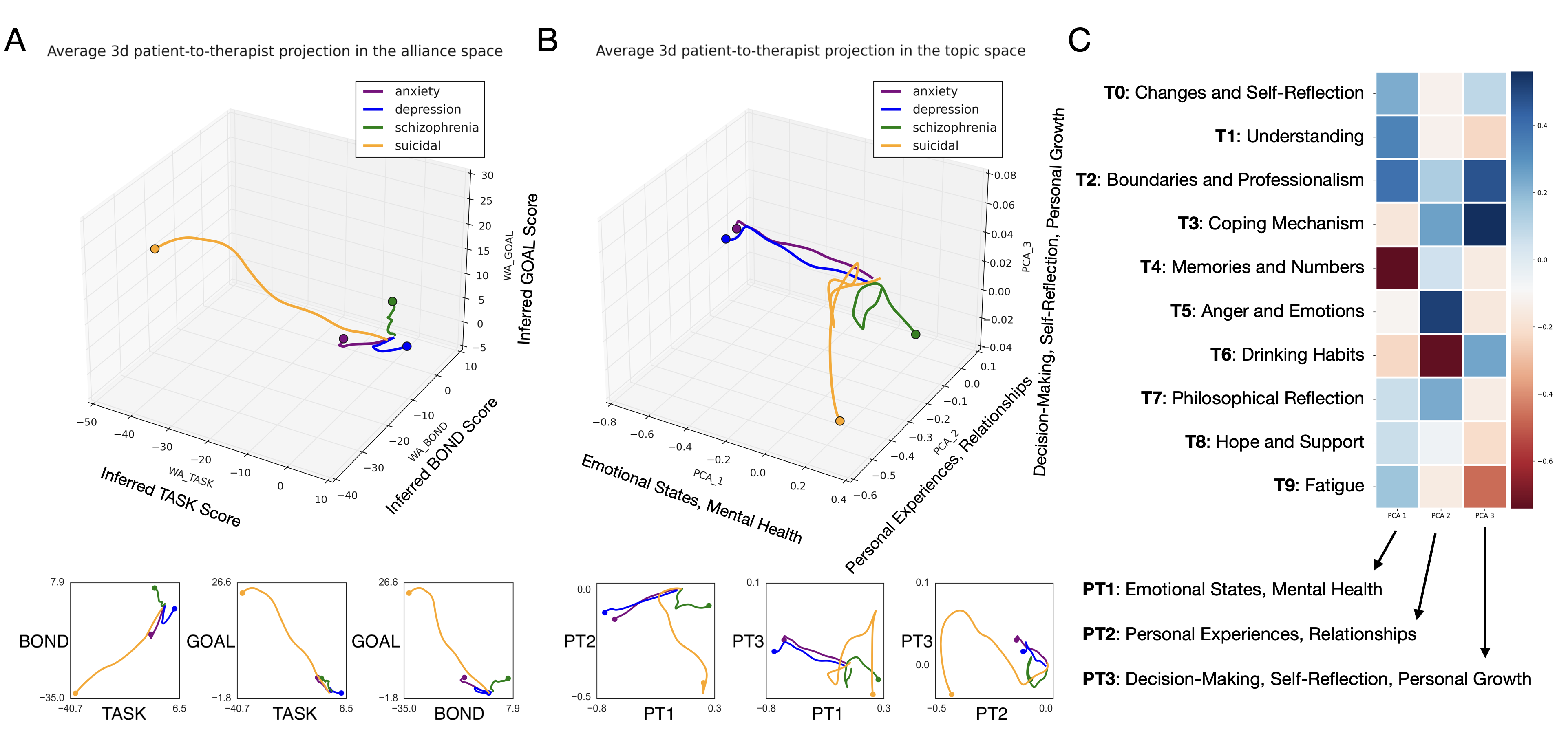} 
\caption{\textbf{The average 3d trajectories of different classes of psychiatric conditions in the alliance and topic space.} For each clinical condition, we averaged the time series of the therapists and patients over the sessions. We compute the patient-therapist discrepancy and their cumulative sum over time, both in terms of their inferred scores of working alliance (A) and topic scores (B). In both the alliance space and the topic space, we mark the end points of the trajectories as a bigger dot. The coefficients of the three principal topics are shown as a heatmap in panel C.
}\label{fig:trajs}
\end{figure}

\subsection{Analyzing the Temporal Dynamics of Topics}

To analyze the temporal dynamics of topics, we compute topic scores at the turn-level. We utilize the Embedded Topic Model (ETM) for this analysis, as it models each word with a categorical distribution based on the inner product between a word embedding and the embedding of its assigned topic \cite{dieng2020topic}. We use the same Word2Vec word embeddings to embed both the topics and the dialogue turns, to then compute the cosine similarity between the embedded topic vector and the embedded turn vector. By applying these methods, which we term Temporal Topic Modeling (TMM), we obtain turn-resolution topic scores that capture the temporal dynamics of the topics discussed during the therapy session. These turn-level topic scores allow us to track the changes in topic relevance over time, providing insights into the progression of the therapy, the emergence of specific topics, and shifts in the focus of the conversation.

As shown in Figure \ref{fig:topic_flowchart}C, we can analyze the association between therapist topics and inferred patient working alliance by annotating both of them at the same time in a turn-level resolution. More specifically, we can estimate how inferred patient working alliance is conditioned by the therapist's choice of topics in their dialogue.

\section{Results}

\subsection{Insights on Patient-Doctor Relationship from Working Alliance Analysis}

In this section, we present the findings from applying working alliance analysis and topic modeling to the psychotherapy dataset.

\textbf{Patient-Therapist Consistency of Working Alliance.}
We investigate the consistency of the working alliance estimation between patients and therapists. Comparing the estimates, we observe that therapists tend to overestimate the working alliance overall. Specifically, therapists tend to overestimate the task and bond scales, but underestimate the goal scale. These differences are statistically significant ($p<0.001$). We also find that the working alliance scores differ significantly between certain pairs of psychiatric conditions, such as anxiety and depression, and anxiety and schizophrenia, in both the therapist and patient versions ($p<0.001$). Furthermore, the working alliance scores for all four scales can significantly detect individuals with suicidality ($p<0.001$). There are also variations among the working alliance scores of each clinical conditions (Tables \ref{stats_wa_task_t}, \ref{stats_wa_task_p}, \ref{stats_wa_bond_t}, \ref{stats_wa_bond_p}, \ref{stats_wa_goal_t}, and \ref{stats_wa_goal_p} in the Supplementary Materials, for statistical differences of the working alliance scores among conditions).

In TASK scale, significant differences were observed between anxiety and depression (p < 1.550e-26), between anxiety and schizophrenia (p < 3.835e-05), and between depression and suicidal tendencies (p < 1.900e-02) in the therapist turns. Similarly, patient turns revealed significant differences across all pairs of conditions, including between anxiety and depression (p < 1.796e-25), and schizophrenia and suicidal tendencies (p < 5.465e-14). These results suggest that therapists may approach task alignment differently depending on the patient’s psychiatric condition. The significant differences observed in both therapist and patient turns imply that therapists might align tasks more effectively with anxious and depressed patients compared to those with schizophrenia or suicidal tendencies.

In BOND scale, The therapist turns showed significant differences in the BOND scale between anxiety and depression (p < 1.600e-22), as well as between anxiety and schizophrenia (p < 2.998e-05). Patient turns revealed significant differences across almost all pairs, including between schizophrenia and depression (p < 3.883e-03), and schizophrenia and suicidal tendencies (p < 3.629e-34). These results suggest that therapists may perceive stronger bonds with anxious and depressed patients compared to those with schizophrenia or suicidal tendencies.

In GOAL scale, significant differences were found across all therapist pairs except between depression and schizophrenia (p = 4.959e-01). In the patient data, significant differences were found only between schizophrenia and anxiety (p < 8.006e-04), and between schizophrenia and depression (p < 4.519e-05). This suggests notable discrepancies in how therapists and patients perceive the alignment of treatment goals across these psychiatric conditions.

These variations are important because they reveal distinct patterns in the therapeutic alliance based on the clinical condition being treated. For example, patients with suicidal tendencies exhibit significantly larger discrepancies in their working alliance scores compared to other conditions, particularly in the TASK and BOND scales, but not as much from the therapist's perspective. However, in the GOAL scale, therapists show significant differences across conditions, while patients do not perceive these differences as strongly. This mismatch may indicate that therapists are more attuned to differences in goal alignment across conditions than patients are, especially when it comes to conditions like anxiety and suicidality. Such discrepancies highlight the challenges therapists face in aligning therapeutic goals with certain patient populations, particularly those with suicidal tendencies, and suggest that more nuanced strategies are needed to ensure that both parties are on the same page regarding treatment goals.

\textbf{Temporal Dynamics of Working Alliance.}
We examine the temporal dynamics of the working alliance by mapping the trajectories in the alliance space for the three major scales (task, bond, and goal). Specifically, we calculate the difference between patient and therapist scores to explore perceived misalignments in these aspects. Figures \ref{fig:avg_score} and \ref{fig:trajs} 
illustrate the average trajectories across different psychiatric conditions, revealing several distinctive patterns.

For suicidality, the trajectory in bond and task scales shows a continuous decrease, moving increasingly negative over time, which suggests a growing misalignment as therapy progresses. This negative trend may reflect a disconnect in emotional rapport and task engagement between patient and therapist for this group. In contrast, the goal scale trajectory for suicidality shifts from zero to positive, implying an improving alignment on long-term goals. This may also reflect the particular challenges and variability in maintaining a strong alliance with suicidal patients, where initial engagement might wane over time as treatment progresses. However, it’s worth noting that our limited sample size for suicidality could introduce bias, so these patterns should be interpreted cautiously.

In the cases of anxiety and depression, we observe that both conditions start with a negative trajectory in the bond scale, which intensifies in the first half of the sessions. However, both conditions show a partial bounce-back in bond alignment later, even though the values remain predominantly negative, indicating a persistent gap in perceived emotional connection. For the task scale, the trajectories diverge: depression sessions increase from zero to positive, indicating improving alignment in task engagement, whereas anxiety sessions drop from zero to negative, suggesting a growing misalignment over time.  This divergence could imply that task-focused interventions may need to be adapted differently for these two conditions over time.

For schizophrenia, the trajectories in both bond and goal scales display a consistent increase, remaining positive throughout the sessions. This suggests that for schizophrenia, patient-therapist alignment improves over time, particularly in terms of emotional connection and goal setting, which could indicate a progressive establishment of rapport and shared understanding in this population.

\begin{table}[tb]
      \caption{ \textbf{Classification accuracy of psychotherapy sessions.} Results represent percentual accuracy; chance accuracy of this 4-class classification task is 25\%, boldface indicates statistical significance at $z$-value $\ge 4.3$ ($p$-value $\le 10^{-5}$). 
      }
      \label{tab:classification} 
      \centering
      \resizebox{\linewidth}{!}{
 \begin{tabular}{ l | c | c | c | c | c | c }
  & \multicolumn{3}{c}{SentenceBERT} \vline & \multicolumn{3}{c}{Doc2Vec} \\
  & Patient turns & Therapist turns & Both turns & Patient turns & Therapist turns & Both turns \\ \hline
WAT (inferred score + pretrained embedding) & 27.6 & 27.0 & \textbf{26.0} & \textbf{34.1} & 25.7 & \textbf{31.9} \\
WAT (inferred score) & 26.1 & 23.4 & 25.5 & 28.9 & 23.7 & \textbf{31.9} \\
Embedding Transformer & 24.8 & 24.0 & 25.5 & \textbf{31.8} & 26.2 & 29.9 \\ \hline
WA-LSTM  (inferred score + pretrained embedding) & \textbf{35.0} & \textbf{36.9} & 23.3 & \textbf{46.0} & 27.7 & 29.6 \\
WA-LSTM (inferred score) & 24.5 & \textbf{34.2} & 22.6 & 30.2 & 24.7 & \textbf{43.4} \\
Embedding LSTM & 23.0 & \textbf{36.0} & 22.9 & \textbf{44.3} & \textbf{31.1} & \textbf{31.1} \\ \hline

\end{tabular}
 }
\end{table}

\subsection{Inference of the Diagnosis of Clinical Conditions}

Next, we evaluate the usefulness of features derived from working alliance and dialogue turns in classifying psychotherapy sessions into four clinical labels. 
This evaluation is intended solely for validation purposes and not as a diagnostic tool. Given that clinical diagnoses were already established and influenced the content of these transcripts, our goal is not to replace diagnostic standards like the Diagnostic and Statistical Manual of Mental Disorders (DSM), but rather to demonstrate that inferred alliance scores can correlate meaningfully with known clinical outcomes. This serves as a way to assess the predictive power of working alliance trajectories within the sessions, suggesting that the inferred scores can provide valuable insights into the types of patients being treated. These findings highlight the potential for working alliance measures to inform and guide treatment strategies based on patterns observable early in therapy sessions.

We employ two classifier backbones: Transformers \cite{vaswani2017attention} and Long Short-Term Memory Networks (LSTM) \cite{hochreiter1997long}. We compare three types of features: inferred scores of working alliance, pretrained document embedding, and both. Additionally, we explore the performance using dialogue turns from patients, therapists, and both patients and therapists.

We address the imbalanced nature of the dataset by using a sampling technique during training. The models are trained for over 50,000 iterations using stochastic gradient descent with a learning rate of 0.001 and momentum of 0.9. We report the performance of diagnosis, in another word, the multi-class classification accuracy among the four clinical conditions as four class labels, as the evaluation metric (Table \ref{tab:classification}).

Overall, we observe that using the combined feature of the inferred scores and pretrained embedding in Transformer and LSTM-based models improves classification performance. Among all the models, the WA-LSTM model with the combined feature using only patient turns achieves the highest classification accuracy (46\%), followed by the WA-LSTM model using only the inferred scores of working alliance with turns from both patients and therapists (43.4\%). These results indicate the advantage of incorporating predicted clinical outcomes in characterizing sessions based on their clinical conditions. Additionally, we find that the inference of therapeutic working alliance using Doc2Vec is more beneficial for modeling patient turns, while SentenceBERT is advantageous for both therapist and patient features.

We further examine the influence of different features and embeddings on classification performance. The combined feature of inferred scores of pretrained document embedding and working alliance consistently outperforms the other feature types in both Transformer and LSTM models. 
When using SentenceBERT for sentence embeddings, we observed a modest improvement in classification performance when training on patient turns alone, as opposed to using both patient and therapist data together. This suggests that combining therapist and patient turns may introduce overlapping features that could interfere with the clarity of the inferred alliance scores. For clinicians, this finding highlights the potential value in separately analyzing patient dialogue, as it might yield more distinct and interpretable signals regarding the working alliance.
The Transformers utilizing the combined feature of inferred scores and pretrained embedding, especially when using Doc2Vec as the sentence embedding, achieve the best performance. 

These results indicate the potential usefulness of inferred therapeutic or psychological state scores in downstream tasks, such as supporting treatment planning and personalization. While this part of the study does not aim to replace clinical diagnosis, the inferred scores could offer additional predictive insights that complement existing clinical frameworks. For instance, in a transdiagnostic setting, these early session indicators might help clinicians understand which types of treatment plans are most suitable, even before a clear diagnosis is fully established. Future investigations could explore other downstream tasks and leverage the attention mechanism of Transformer blocks for further interpretation.

\subsection{Topic Modeling Approach to Provide Interpretable Clinical Insights}

We apply five state-of-the-art neural topic modeling approaches to the psychotherapy dataset and analyze the learned topics. We divide the transcript sessions into three categories based on psychiatric conditions (anxiety, depression, and schizophrenia) and train topic models for each category. We evaluate the models using various coherence and diversity metrics to assess the quality of the generated topics.

\begin{table}[!h]
\caption{{Therapeutic Topics and Interventions: topics description and likely interventional rationale. }}\label{tab:topic_interpretations}
      \centering
      \resizebox{\linewidth}{!}{
\begin{tabular}{l p{0.18\linewidth}  p{0.7\linewidth} p{0.15\linewidth}}
\hline
& \textbf{Topics} & \textbf{Interventions} & \textbf{Examples} \\
\hline
 &  {\vfill \textbf{Topic 0}: Encouraging Change and Self-Reflection \vfill} & 
 \begin{itemize}[topsep=0pt,leftmargin=*]
\item Explore the potential for the individual to reconsider their viewpoints or habits.
\item Suggest maintaining certain behaviors for a period of time.
\item Discuss personal growth and maintaining a sense of self-awareness.
\end{itemize} 
 &
 {\vfill ``Well, I mean well keep doing the exercise.'' \vfill} \\
\hline
&  {\vfill \textbf{Topic 1}: Making a Case and Seeking Understanding \vfill} & \begin{itemize}[topsep=0pt,leftmargin=*]
\item Discuss the importance of presenting a clear argument or perspective.
\item Inquire about interactions with a case manager or authority figure.
\item Explore the depth of personal experiences and the case's significance in the person's life.
\end{itemize} 
 &
 {\vfill ``You've made your case that weight's important.'' \vfill} \\
\hline
&  {\vfill \textbf{Topic 2}: Maintaining Boundaries and Professionalism \vfill} & 
 \begin{itemize}[topsep=0pt,leftmargin=*]
\item Discuss the importance of establishing emotional boundaries.
\item Encourage the person to continue engaging in therapeutic exercises.
\item Reflect on maintaining professionalism and boundaries within the therapeutic relationship.
\end{itemize} 
&
 {\vfill ``Yes, you keep it very professional; patient-doctor.'' \vfill} \\
\hline
 &  {\vfill \textbf{Topic 3}: Discussing Coping Mechanisms and Playfulness \vfill} & 
 \begin{itemize}[topsep=0pt,leftmargin=*]
\item Encourage the person to engage in activities that bring them joy.
\item Explore social interactions and relationships, such as playdates and friendships.
\item Discuss using adaptive strategies like ``playing it by ear'' to navigate life's uncertainties.
\end{itemize} 
 &
 {\vfill ``Play it up.'' \vfill} \\
\hline
 &  {\vfill \textbf{Topic 4}: Focus on Specific Numbers and Memories \vfill} & 
 \begin{itemize}[topsep=0pt,leftmargin=*]
\item Discuss specific instances involving numbers (e.g., taking medication).
\item Explore memories associated with certain days or events.
\item Engage in conversation related to quantifiable aspects of the person's life.
\end{itemize} 
&
 {\vfill ``Just for like 1 day? Okay, how do you remember that day?'' \vfill} \\
\hline
 &  {\vfill \textbf{Topic 5}: Exploring Anger and Emotions \vfill} & 
 \begin{itemize}[topsep=0pt,leftmargin=*]
\item Inquire about feelings of anger towards specific individuals.
\item Validate and explore the intensity of anger towards others.
\item Investigate triggers, circumstances, and potential underlying emotions contributing to anger.
\end{itemize} 
&
 {\vfill ``Are you angry with him at all?'' \vfill} \\
\hline
&  {\vfill \textbf{Topic 6}: Discussing Drinking Habits \vfill} & 
 \begin{itemize}[topsep=0pt,leftmargin=*]
\item  Inquire about the individual's alcohol consumption.
\item Explore preferences related to different beverages, like soda and coffee.
\item Address drinking habits and patterns to gain insights into lifestyle choices.
\end{itemize} 
 &
 {\vfill ``Do you drink any, or almost none?'' \vfill} \\
\hline
 &  {\vfill \textbf{Topic 7}: Philosophical Reflections and Communication \vfill} & 
 \begin{itemize}[topsep=0pt,leftmargin=*]
\item Engage in discussions about human nature and existence.
\item Reflect on personal experiences and projects.
\item Explore communication dynamics, including misperceptions and honesty.
\end{itemize} 
 &
 {\vfill ``Every human being?'' \vfill} \\
\hline
&  {\vfill \textbf{Topic 8}: Offering Hope and Support \vfill} & 
 \begin{itemize}[topsep=0pt,leftmargin=*]
\item Express empathy and hope for positive outcomes.
\item Acknowledge the individual's progress and insights.
\item Provide encouragement to continue the journey of self-discovery and growth.
\end{itemize} &
{\vfill ``I hope so too. Take care.'' \vfill} \\
\hline
&  {\vfill \textbf{Topic 9}: Addressing Fatigue \vfill} & 
 \begin{itemize}[topsep=0pt,leftmargin=*]
     \item Inquire about the reasons behind the individual feeling tired.
     \item Explore the decision-making process that led to seeking therapy.
     \item Elicit the motivations and factors that brought the person to the therapy session.
 \end{itemize} &
 {\vfill ``Tired a lot?'' \vfill} \\
\hline
\end{tabular}
}
\end{table}

\clearpage

The evaluation metrics include topic embedding coherence ($c_v$, $c_{w2v}$, $c_{uci}$, $c_{npmi}$) and other measures such as coherence based on the asymmetrical confirmation measure and topic diversity. The results show variability in the rankings across the coherence metrics (Table \ref{tab:topic_eval} and \ref{tab:emb_eval} in the Supplementary Materials), but certain models consistently demonstrate higher topic coherence and diversity (for both metrics, the higher, the better). Following \cite{mimno2011optimizing,dieng2020topic}, the topic diversity is computed as the proportion of unique words (PUW) with 1.0 being perfectly diverse, and the topic coherence is computed as an integrated log ratio between the co-document frequency and document frequency with higher being better (as correlated by expert ratings). Notably, the Wasserstein-based Topic Models and Embedded Topic Models exhibit relatively high coherence and diversity in these two metrics.

\subsection{Major Themes in Psychotherapeutic Topics and Interventions}

To gain further insights, we provide topic interpretations by examining the highest scoring turns associated with each topic. The interpretations reveal the dominant themes in the dialogue for different topics and provide insights into patients' emotional states, personal experiences, and self-reflection. In the context of performing topic modeling on the text corpus of the entire psychotherapy dataset, the goal is to identify the top 10 topics and extract more distinctive features for downstream tasks. We perform the topic modeling on the entire text corpus to maintain the coherence within the patient-therapist dialogue, but are interested in the strategies and themes of the therapists in their contributions within the contexts of these learned topics. To achieve this goal, we ranked the therapist's dialogue turns by their topic scores (the higher the score is, the more likely it was related to a particular topic), and then picked out the top 10 sentences for each topic as exemplar ones. 

To expand interpretability possibilities, and diminish the effect of our biases, we resorted to a generative Large Language Model (LLM), ChatGPT based on GPT-3.5, and prompted it for summaries of the discussed topics as follows:

\textit{``I have the following top sentences exemplifying ten topics. Can you summarize what the three interventions items attributed to each topic spaces the therapists are talking about, respectively? For instance, what therapeutic intervention the therapist is applying.'' 
}

The result of this analysis is presented in Table \ref{tab:topic_interpretations}, which lists the high-level description of each topic, the likely interventional rationale and a literal example. We found that the vast implications of these topics and interventions can be partially summarized into four major themes with the prompt \textit{``can you summarize them into a few major themes?''}, as follows:

\textbf{Theme 1: Self-Exploration and Personal Growth.}
Therapists engage clients in discussions centered around self-discovery, personal transformation, and introspection. Within this theme, the ``Changes and Self-Reflection'' (Topic 0) topic encourages individuals to reconsider viewpoints, maintain specific behaviors, and explore personal growth opportunities. ``Memories and Numbers'' (Topic 4) delves into specific instances tied to numbers and memories, aiding clients in recalling significant life events. ``Hope and Support'' (Topic 8) contributes to optimism and resilience by expressing empathy, acknowledging progress, and providing encouragement. Lastly, addressing ``Fatigue'' (Topic 9) involves understanding reasons for tiredness, exploring motivations behind seeking therapy, and examining factors influencing emotional and physical well-being.

\textbf{Theme 2: Understanding and Communication.}
This theme revolves around effective communication, mutual understanding, and deep introspection. In the context of ``Understanding'' (Topic 1), therapists guide clients in presenting their viewpoints coherently and exploring the significance of personal experiences. This dovetails with ``Philosophical Reflection'' (Topic 7), where therapists engage in discussions about human nature, existence, and personal projects. These topics emphasize the importance of effective communication and introspective exploration for therapeutic progress.

\textbf{Theme 3: Emotional Well-being and Coping Strategies.}
Addressing emotional well-being is at the core of this theme, with a focus on coping mechanisms, anger, and adaptive strategies. ``Coping Mechanism'' (Topic 3) involves encouraging clients to engage in activities that bring joy, navigate uncertainties, and manage stress. ``Anger and Emotions'' (Topic 5) explores feelings of anger, validates emotional intensity, and investigates underlying triggers. ``Drinking Habits'' (Topic 6) allows therapists to delve into substance-related behaviors, uncovering potential unhealthy coping mechanisms and aiding clients in making informed choices.

\textbf{Theme 4: Therapeutic Relationship and Ethical Boundaries.}
This theme revolves around maintaining a healthy therapeutic relationship through the establishment of boundaries and professionalism. ``Boundaries and Professionalism'' (Topic 2) highlights the importance of emotional boundaries, encourages clients' engagement in therapeutic exercises, and underscores the significance of professionalism in the therapist-client relationship.

The \textit{major themes} identified through this topic modeling approach represent the key therapeutic strategies and intervention techniques utilized by the therapist during sessions. These themes, directly related to intervention strategies such as cognitive reframing, emotional validation, or goal-setting, reflect the therapist’s active role in steering the therapeutic process. By focusing on the therapist's contributions, we can better understand not just the content of the conversations, but also the clinical strategies that are emphasized in practice. These themes are particularly useful for extracting actionable insights that clinicians might find helpful in refining their approaches to treatment.

The findings from our topic modeling analysis reveal several major themes that underpin the diverse array of therapeutic interactions observed in psychotherapy sessions. These themes provide a comprehensive framework for understanding the nuanced ways therapists guide clients through self-exploration, emotional processing, communication enhancement, and personal growth journeys.

In contrast, the \textit{principal component topics}, which are discussed in our next section, provide a broader view of the full dialogue between patient and therapist. These topics are learned from both therapist and patient turns and offer a general understanding of what is being talked about throughout the sessions, serving as a comprehensive reference for the entire discourse. Together, the major themes and principal component topics complement one another: the themes give us insight into the specific therapeutic techniques applied, while the topics provide a broader contextual framework for the overall content of therapy.

\subsection{Principal Topic Space in Psychotherapy Sessions}

To further extract more distinctive features from the 10 topics for downstream tasks, a principal component analysis is performed on the topic space. This analysis enables the identification of three principal topic spaces that encompass the patient turns and the corresponding therapeutic interventions taken by the therapists. The coefficients of the principal components are presented at Figure \ref{fig:trajs}C.

To expand interpretability possibilities, and diminish the effect of our biases, we resorted to a generative Large Language Model (LLM), ChatGPT based on GPT-3.5, and prompted it for summaries of the principal topics as follows:

\textit{``I have the following top sentences exemplifying three principal topic spaces. Can you summarize what the three topics the patients are talking about, respectively?''}, and \textit{``Again, I have the following top sentences exemplifying the three principal topic spaces. Can you summarize what the three intervention items attributed to each principal topic spaces the therapists are talking about, respectively? For instance, what therapeutic intervention is the therapist applying.''} 

In the following subsections we present the interpretation result of principal topics for patients and therapists.

\textbf{Principal Component Topic 1: Emotional States and Mental Health.} 
The first principal component topic revolves around patients' emotional states and mental health. It encompasses discussions about emotions, mood, and mental well-being. Patients often express their feelings of depression, anger, anxiety, and powerlessness. Therapists respond by providing empathy, validation, and encouragement to help patients navigate their emotions. The interventions attributed to this topic include:

\begin{enumerate}
\item Validation and Empathy: Therapists acknowledge and validate patients' emotions, creating a safe space for them to express their feelings without judgment.
\item Encouragement and Support: Therapists motivate patients to continue their progress and efforts, emphasizing the importance of self-care, well-being, and healthy routines.
\item Exploration and Understanding: Therapists engage patients in exploring their emotions and thoughts, helping them gain insights into their experiences and develop strategies for coping and personal growth.
\end{enumerate}

\textbf{Principal Component Topic 2: Personal Experiences and Relationships}
The second principal component topic centers around patients' personal experiences and their relationships with others. It encompasses discussions about family dynamics, past experiences, and interpersonal relationships. Patients may express anger, sadness, or difficulties in their relationships. Therapists address these emotions and provide guidance for navigating relationships. The interventions attributed to this topic include:

\begin{enumerate}
\item Relationship Exploration: Therapists help patients explore their relationship dynamics, understand their emotions, and develop healthier ways of relating to others.
\item Validation and Support: Therapists validate patients' experiences and provide support during challenging relationship situations, creating a space for reflection and growth.
\item Communication and Conflict Resolution: Therapists assist patients in improving communication skills, conflict resolution, and establishing boundaries to enhance their interpersonal relationships.
\end{enumerate}

\textbf{Principal Component Topic 3: Decision-Making, Self-Reflection and Personal Growth}
The third principal component topic focuses on self-reflection and personal growth. It encompasses discussions about self-perception, beliefs, and aspirations for personal development. Patients often express desires for change, self-improvement, and gaining a deeper understanding of themselves. Therapists foster self-reflection and guide patients towards personal growth. The interventions attributed to this topic include:

\begin{enumerate}
\item Self-Reflection and Insight: Therapists encourage patients to engage in self-reflection, explore their beliefs, values, and aspirations, and gain deeper insights into themselves.
\item Goal Setting and Planning: Therapists collaborate with patients to set meaningful goals, develop strategies, and create intervention plans to support personal growth and progress.
\item Empowerment and Skills Development: Therapists empower patients by helping them identify their strengths, build resilience, and acquire coping skills to navigate challenges and achieve personal growth.
\end{enumerate}

\textbf{Variability Among Different Clinical Conditions.}
Our dataset comprises four subsets, each representing a distinct patient population characterized by a specific clinical condition. To capture this variability, we trained individual topic models for each subpopulation cohort. The learned topics were analyzed to highlight condition-specific themes and insights.

In the sessions with anxiety patients, therapists may place particular emphasis on managing anxiety symptoms, addressing fears and worries, and implementing coping strategies for anxiety management. The interventions attributed to therapists in this dataset may include techniques such as relaxation exercises, cognitive restructuring, and exposure therapy.

In the sessions with depression patients, therapists may focus on understanding and alleviating symptoms of depression, exploring the underlying causes of depressive feelings, and promoting self-care and self-compassion. The interventions attributed to therapists in this dataset may involve behavioral activation, cognitive reframing, and facilitating social support systems.

In the sessions with schizophrenia patients, therapists in this dataset may prioritize addressing symptoms related to psychosis, managing hallucinations or delusions, and enhancing reality testing and medication adherence. The interventions attributed to therapists may involve psychoeducation about the illness, cognitive-behavioral interventions for managing symptoms, and collaborating with other healthcare providers.

In the sessions with suicidal patients, and therapists in this dataset may prioritize risk assessment, crisis intervention, and safety planning. The interventions attributed to therapists may involve creating a supportive and non-judgmental environment, assessing suicidal ideation and intent, and implementing strategies to reduce immediate risk while developing long-term coping skills.

\textbf{Temporal Dynamics of Therapy Topics.}
We further investigate the temporal dynamics of the learned Embedded Topic Model, which yields the best performance, on the text corpus of the entire psychotherapy dataset. By computing topic scores for each turn, we can analyze the dialogue dynamics within the topic space. Figure \ref{fig:trajs} presents the cumulative discrepancy of scores in the three scales (panel A) and of projections on the three principal topics (panel B) between patient-therapist dialogue pairs across topics. 


Our analysis reveals several distinctive patterns. Specifically, we observe the largest growing discrepancy for suicidality, similarities in topic divergence between anxiety and depression, and a complementary pattern for schizophrenia, which is most prominent in principal topic 1 (PT1) related to emotional states and mental health.

For anxiety and depression, trajectories are more separable, with both conditions showing a consistent decline from zero to negative values in PT1 (emotional states and mental health), suggesting an increasing divergence between patients and therapists on these issues. In PT2 (personal experiences and relationships), the discrepancy also becomes gradually more negative, while in PT3 (decision-making and personal growth), both conditions show a shift from zero to positive. This indicates growing alignment in discussing personal growth, despite increasing misalignment on emotional and relational topics.

For schizophrenia, we observe a complementary pattern where PT2 shows an initial decline from zero to negative but stabilizes midway, suggesting early divergence in personal relationships. Meanwhile, PT1 shifts positively, starting after PT2 stabilizes, creating a twist in the trajectory. Additionally, PT3 initially dips negative before rebounding to positive. This evolving pattern implies that for schizophrenia, therapists and patients might initially diverge on personal experiences, but increasingly align on emotional health and self-reflection as sessions progress.

Suicidality shows the widest range of topic divergence. The trajectory for PT2 begins with a drop in alignment, indicating early divergence in discussions about personal experiences. PT3, meanwhile, initially rises from zero to positive but then sharply reverses to negative, suggesting fluctuating alignment on decision-making and personal growth topics. This spread-out pattern may reflect the more dynamic and complex nature of therapeutic engagement with suicidal patients, which can involve rapid shifts in focus.

We also observe that trajectories are generally more separable in anxiety and schizophrenia sessions, while depression sessions tend to be more entangled across topics. This separability may indicate distinct thematic focus patterns, which could offer therapists insights into how patients engage with different therapeutic areas based on their conditions.

 \begin{figure}[tb]
\centering
\includegraphics[width=\linewidth]{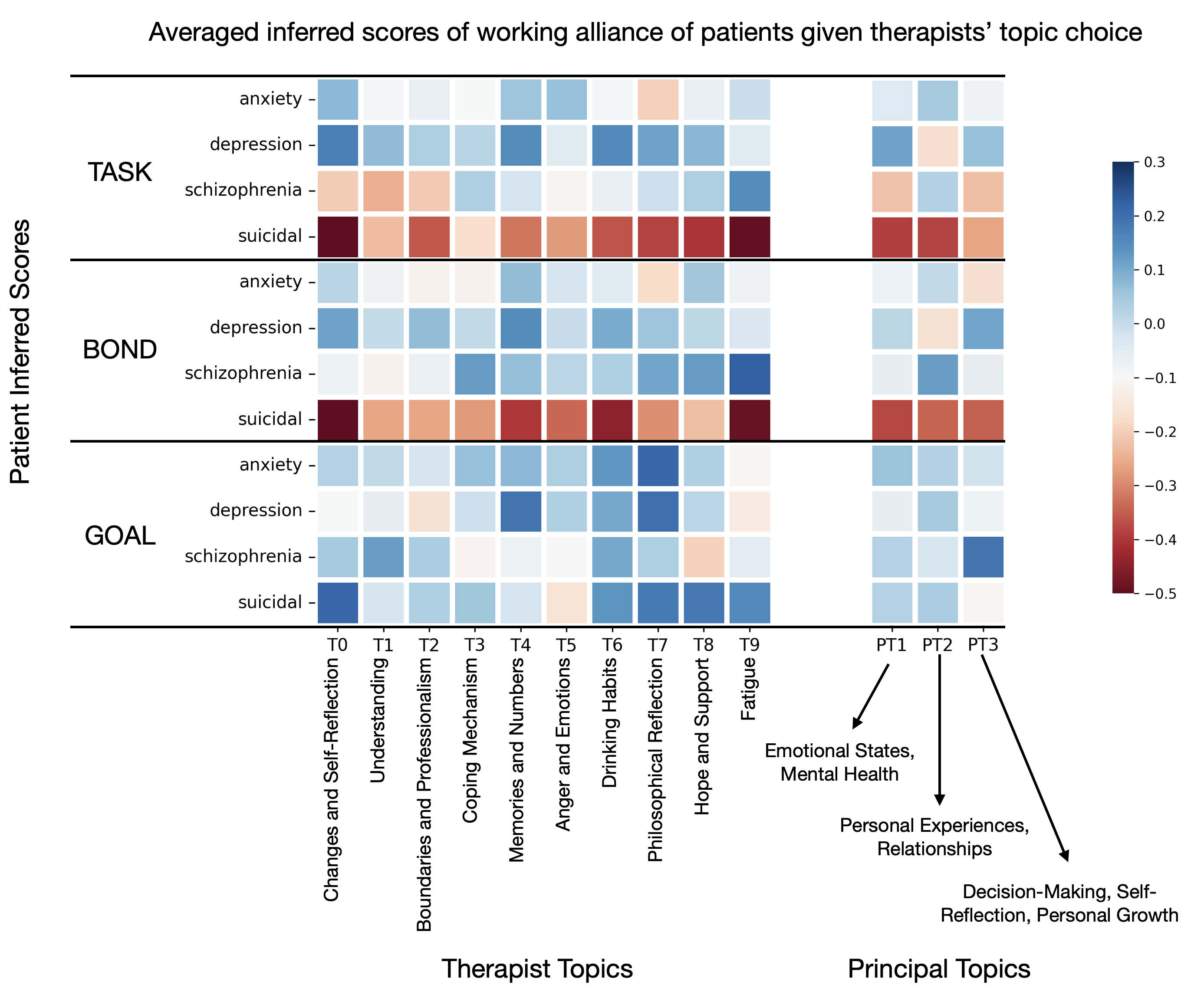}
\caption{\textbf{Patients' working alliance differ when therapists chose different topics to discuss.} We compute the topic (T) and principal topic (PT) scores for all the therapist turns, and select the top 100 turns for each clinical condition and each topic; to estimate the effect of the topic on working alliance, we compute the average working alliance scores of the subsequent patients' turns. We plot these averaged working alliance scores by the patients in a heatmap.
}\label{fig:topic_wa_filter}
\end{figure}

\textbf{Topic-Alliance Associations as Actionable Insights for Clinicians.}
To explore the informative value of topics for therapeutic insights, we combine topic modeling with the inferred working alliance (Figure \ref{fig:topic_wa_filter}). By filtering the therapist turns with high topic scores, we plot the average working alliance scores for corresponding patient turns. We observe distinctions among the effects on patients' working alliance across different topics and clinical conditions. For example, discussing tiredness and decision-making positively influences the bond and task scales in schizophrenia patients but has less impact on other patients. Additionally, discussing sickness, self-injuries, and coping mechanisms positively affects the task scale in depression patients and the goal scale in suicidal patients. 

If the clinicians discuss principal component topic 1, ``Emotional States and Mental Health'', it increases the TASK and BOND scales for depression patients, but decreases them for suicidal patients. 

If the clinicians discuss principal component topic 2, ``Personal Experiences and Relationships'', it increases GOAL scale for all patients, helps with the BOND scale for schizophrenia patients, but decreases the TASK and BOND scales for depression and suicidal patients.

If the clinicians discuss principal component topic 3, ``Decision-Making, Self-Reflection and Personal Growth'', it might increase the BOND scale for depression patients and GOAL scale for schizophrenia patients, but decrease the BOND and TASK scales for suicidal patents. 

As a section summary, the application of topic modeling to psychotherapy transcripts offers interpretable insights into the dominant themes and dynamics of therapeutic dialogues. It enables the identification of key topics related to emotional states, personal experiences, and self-reflection. Combined with the analysis of inferred working alliance, this approach provides valuable information for understanding the therapeutic process and potentially highlighting topics and dialogue segments indicative of therapeutic breakthroughs.

\section{Discussion}

\subsection{Working Alliance Analysis}

Our analysis of the psychotherapy transcripts provides valuable insights into the working alliance between therapists and patients. We observe systematic differences in the mean inferred alliance scores between patients and therapists, as well as variations across different psychiatric disorders. However, the analysis of the in-session evolution of the working alliance scores reveals more interesting dynamics.

In particular, we find that while all conditions show a systematic misalignment of scores between patients and therapists, this misalignment is significantly more pronounced for suicidality. This observation is evident in both the mean scores and the temporal trace of the full and sub-scales. In contrast, anxiety and depression display a clear trend for convergence in the full and bond scales as the therapy sessions progress, which is not observed in the task and goal scales, nor in schizophrenia or suicidality. These features of the therapeutic dialogue, such as the alignment and convergence of scores, can provide valuable insights into the therapeutic process and have implications for diagnosing and treating neuropsychiatric conditions \cite{Janna2020}.

The analysis of past therapy sessions, as well as real-time sessions, has the potential to help therapists identify key segments of therapy leading to breakthroughs. By leveraging computational modeling and statistical optimization, therapists can compound their expertise with causal and predictive analytic modeling to enhance their understanding of the therapeutic process. Additionally, trainees can benefit from studying annotated versions of sessions conducted by experts, further sharpening their intuition and skills. Furthermore, the integration of generative language models and statistical optimization techniques may enable the design of limited chatbots for triage and emergency response in mental health care \cite{Garg2020}.

\textbf{Clinical Implications and Roadmap.}
These distinct trajectories offer a preliminary roadmap for clinicians. By identifying prototypical alliance trajectories for each condition, therapists can begin to map individual sessions onto these broader patterns. For example, if a session with a suicidal patient diverges significantly from the typical trajectory (e.g., a sharper decline in the bond scale), this might signal a need for immediate intervention.

These patterns of alignment and misalignment provide clinicians with a roadmap for interpreting patient-therapist dynamics over time. Furthermore, therapists could use these trajectories to inform treatment adjustments, with specific attention to each scale:

In BOND scale, a downward trend in bond from the patient's perspective, especially with suicidal or anxious patients, could prompt a focus on strengthening rapport and emotional connection. For suicidality, the continuous decline in bond alignment suggests an increasing emotional disconnect. Clinicians might focus on interventions that specifically aim to build rapport with suicidal patients, addressing the emotional aspects of therapy to prevent further drift.

In TASK scale, the divergence in task alignment between anxiety and depression implies that therapists might need to adapt task-related strategies differently for these conditions. The decreasing task alignment for anxiety, moving towards negative values, suggests re-evaluating the relevance of tasks to improve engagement, while the positive trend in depression could indicate opportunities to reinforce shared tasks.

In GOAL scale, the increasing goal alignment for schizophrenia, which remains positive, suggests that long-term goal alignment is progressively achieved, potentially validating goal-oriented therapeutic approaches. Conversely, the improving goal alignment in suicidality may signal a rare but valuable area of common ground between patient and therapist, which could be leveraged to foster hope and engagement.

Mapping individual sessions onto these trajectories could allow therapists to identify early warning signs of divergence from these patterns. For instance, observing a sudden drop in task or bond alignment could prompt a re-evaluation of the therapeutic approach for patients with similar conditions. Future work could focus on developing tools to track these alignment differences in real-time, enabling clinicians to make dynamic adjustments based on evolving alliance data.

\textbf{Limitations.}
While our approach proves effective, there are limitations inherent in inferring psychological states from text data. One potential limitation is the reliance on semantic similarity measures, which may not fully capture the directionality of similarity. For example, it may not differentiate between a statement that is irrelevant to the inventory item and a statement that is opposite to the inventory item. To address this, alternative approaches can incorporate counter arguments and compute similarity scores based on the difference between positive and negative similarity values. However, our findings suggest that the sentence embedding we use already captures the concept of negation, and we observe no clear difference in performance between the alternative approach and our prototypical approach.

Furthermore, our research faces the challenge of limited clinical validation in this field. While the Working Alliance Inventory (WAI) is a widely used measure for assessing working alliance,  it is not a standard practice across all therapeutic settings and does not represent a comprehensive gold standard, particularly for granular, turn-by-turn analyses. Current methods rely on approximate measures and operational approaches that cannot fully capture the nuanced, real-time dynamics of therapeutic interactions. In future work, clinical studies incorporating detailed, turn-by-turn evaluations by human annotators, albeit challenging and resource-intensive, may provide more direct validation of inferred alliance scores within real-world settings and intervention contexts.

Lastly, it is important to note that the dataset had limited access to sessions involving suicidal patients, which might impact the robustness of our findings related to this group. 

\subsection{Topic Modeling and Interpretability}

Topic modeling of the psychotherapy transcripts offers interpretable insights into the content and themes of the therapeutic dialogue. By training topic models on the text corpus of each psychiatric condition and evaluating them using coherence and diversity measures, we obtain meaningful topics associated with anxiety, depression, schizophrenia, and suicidal patients. These topics capture the prevalent themes and concerns within each condition.

For example, in anxiety sessions, topics encompass discussions about fears, worries, coping strategies, and relaxation exercises. In depression sessions, topics revolve around self-esteem, mood, relationships, and social support. In schizophrenia sessions, topics include family dynamics, symptoms of psychosis, and coping strategies. In suicidal sessions, topics may focus on risk assessment, crisis intervention, and safety planning.

The analysis of the temporal dynamics within the learned topics provides further insights into the therapeutic process. By mapping the topic scores of dialogue turns to the working alliance scores, we can identify topics and dialogue segments that are potentially indicative of therapeutic breakthroughs. For instance, in depression sessions, the topic related to self-esteem may be associated with improvements in the bond and task scales of the working alliance. Similarly, in sessions with patients with schizophrenia, the topic related to family dynamics may contribute to positive changes in the bond scale.

\textbf{Clinical Implications and Roadmap.}
By visualizing these trajectories, therapists can retrospectively examine session dynamics, providing an automatic annotation and notetaking mechanism. For junior therapists, this can serve as a valuable tool for skills development, allowing them to understand patient-therapist interactions and adjust strategies based on alignment patterns. Moreover, these insights could guide therapists in real-time, particularly when deviations from expected trajectories are observed. For example:

In PT1, the increasing divergence for anxiety and depression suggests that ongoing attention to emotional topics may be necessary to realign patient-therapist perceptions. 

In PT2, early divergence in schizophrenia and suicidality indicates that a focus on personal relationships could be essential for establishing common ground.

In PT3, the positive alignment trajectory in schizophrenia and the erratic patterns in suicidality suggest that therapists might need to dynamically adjust strategies around personal growth and decision-making.

Future applications could involve real-time tracking of these discrepancies, enabling therapists to monitor and respond to evolving alignment patterns within sessions. This dynamic feedback could improve treatment outcomes by helping clinicians address emerging alignment gaps early in the therapeutic process.

\textbf{Limitations.}
However, it is important to note that the learned topics may exhibit variability across different datasets due to the specific patient populations represented. Each psychiatric condition has its unique challenges and therapeutic goals, which require tailored approaches and interventions by the therapists. Therefore, the topics identified within each condition reflect the prevalent themes and concerns specific to that condition.

One of the strengths of our topic modeling approach is its interpretability. By analyzing the top scoring turns within each topic, we can gain a deeper understanding of the concepts and discussions underlying the topics. This enables clinicians and researchers to explore specific aspects of the therapeutic process and identify areas of focus for further investigation and intervention.

Moving forward, there are several potential directions for future research. Predicting topic scores as states and training LLM- based chatbots through a human-in-the-loop reinforcement learning mechanism based on these states could enhance the capabilities of AI in mental health care to better align with clinical purposes. For clinical practice, an AI knowledge management system that integrates various NLP annotations in real time can be a useful tool \cite{knowledge}. Additionally, studying the relationship between topic modeling and other inference anchors, such as sentiment analysis or linguistic style, could provide a more comprehensive understanding of the therapeutic process in multiple intervention dimensions. 

While these results are informative to a certain degree, we would like to acknowledge the limitations to the methodologies presented. The data we used to train the topic models do not necessarily represent different clinical conditions in a balanced way. In the classification validation task, we have imposed iterative sampling to tackle the imbalance issue, but for training the topic models, we used the full text corpus available, as certain clinical labels (e.g. suicidal patients) only have a handful of sessions available. Due to the limited access to therapy sessions with suicidal patents, the topics characterized by the topic models might be less representative to the conversations happening in this particular patient groups. The Alexander Street dataset mentions that they have more clinical conditions other than the analyzed 4 classes, but due to access limitations, we can only obtain the 4 classes we presented. As open science and data sharing initiatives in the psychiatry domains become more prominent, we believe our methodologies can be adapted in a responsible way to a broader spectrum of clinical conditions.

Other than data-related limitations, there are multiple model-specific decisions we applied to train and test our machine learning models. For the validation task, we train and test our sequence classification models to only take the first 50 turns of dialogues, because the length of the sessions vary from 50 to 400 turns. For the analytics and visualization of trajectories, we choose the first 100 turns of each session to be averaged, as any length beyond it can be too variable to interpret in a safe way.

A final limitation is the lack of validation of the interpretation results we obtained using LLMs. A proper validation by human experts is beyond the scope of the present study, and would require a significant effort of marshaling human resources and designing a systematic approach that can accommodate the expected increase in LLM capabilities. We hope that our study can contribute to initiate such effort.

\subsection{Ethical Considerations}

In this section, we address the ethical considerations associated with our work in analyzing psychotherapy transcripts and utilizing AI in mental health care.

First and foremost, our approach was designed fundamentally as a companion tool to provide valuable insights and support to both experienced therapists and junior clinicians, particularly in the educational setting, where the interpretability of our models can inform the strategies employed by seasoned therapists and aid in the training of future mental health professionals.

When working with patient data, privacy and security are of utmost importance. We have followed ethical guidelines and operational suggestions \cite{matthews2017stories,lin2022ethics,graham2019artificial} to ensure the proper anonymization and protection of sensitive information. The dataset we analyzed had all personally identifiable information, such as metadata, user names, identifiers, and doctors' names, throughly removed. In the context of mental health and psychological well-being, there are additional ethical considerations. The emergence of wearable devices, digital health records, brain imaging measurements, smartphone applications, and social media has transformed the landscape of monitoring and treating mental health conditions. However, it is important to approach these advances with caution, as many of these technologies are still in the proof-of-concept stage \cite{graham2019artificial}. Rigorous clinical validation and regulatory approval are necessary before deploying these technologies for patient care and therapeutic decision-making. Machine learning solutions in psychiatry, including our approach, face difficulties in conducting systematic clinical validation and ensuring the generalizability of results \cite{iniesta2016machine}. Real-world applications often involve small sample sizes, missing data points, and highly correlated variables, which can impact the generalizability and reliability of machine learning models. 

To ensure responsible and safe deployment of AI systems in mental health care, it is necessary to be mindful of potential biases and ethical challenges. Gender bias, language-related ambiguity, and ethnicity-related mental illness connections are examples of such challenges \cite{chen2019can}. Practitioners and machine learning researchers must be aware of these issues and take steps to mitigate biases and promote fairness in AI systems. Engaging the public in discussions about the usage of AI in mental health care is important to foster awareness and avoid unrealistic expectations of AI as a ``domain expert'' \cite{carr2020ai}. We utilized a dataset of over 950 psychotherapy transcripts. While it is the largest dataset available in this research domain, we acknowledge the limitations in terms of its representativeness and generalizability to all populations. The anonymized nature of the dataset and the lack of detailed information about the collection process and demographics of the participants pose constraints. However, we believe that the insights gained from our interpretable investigations are unlikely to increase unforeseeable risks to the patients and have the potential to be valuable in clinical practice.

\section{Conclusions}

In this study, we have introduced an approach that combines state-of-the-art language modeling with therapy-evaluation inventories to provide a detailed representation of the interaction between patients and therapists. Our method offers granular insights for post-session interpretations and has the potential to assist in diagnosing patients based on linguistic features. This dynamic analysis adds a layer of interpretability to the traditional session-based WAI evaluation, and enables us to observe session trajectories and their separability between patients and therapists. We have noted, for instance, that in sessions with patients with anxiety and depression, the trajectories of patients and therapists tend to be more separable, whereas in schizophrenia sessions, they are more entangled. This initial step toward turn-level resolution temporal analysis in topic modeling provides valuable insights that can help therapists improve the effectiveness of psychotherapy. Moreove, while our focus has been on the Working Alliance Inventory, our approach is generic and can be extended to other assessment instruments in the field of psychotherapy.

In conclusion, our analytic framework offers interpretable insights for therapists. By leveraging language models and incorporating temporal analysis, we aim to enhance the understanding of the therapeutic process and support therapists in providing more effective and personalized care to their patients. 

\subsection*{Acknowledgments}

The authors thank Drs. Barnaby Nelson, Alison Yung, and Ravi Tejwani for their helpful discussions and suggestions. We also sincerely appreciate the anonymous reviewers and the editorial team for their constructive feedback and insightful suggestions, which have helped improve this manuscript. Additionally, we acknowledge our collaborators and colleagues for their valuable input and support throughout this study. BL is supported by the Hasso Plattner Foundation, the Windreich Family Foundation, and the National Institutes of Health (NIH). This work was supported in part by the NIH IMPACT-MH PREDiCTOR grant 1U01MH136535-01, which provided funding for BL, YL, RJ, CC, and GC.

\subsection*{Author Contributions} 

BL initiated the study, designed the research, conducted the analysis, wrote the original manuscript, and obtained funding. GC contributed to research design and provided guidance on manuscript writing. CC, YL, and DB provided advice on the study and manuscript writing. All authors reviewed, edited, contributed to, and approved the final manuscript.

\subsection*{Conflict of Interest Statement}

DB and GC are employees of IBM. All other authors declare no competing interests.

\subsection*{Data Availability Statement}

The data analyzed in this study are publicly available from previously published sources. Further details and access information are provided in the references or available upon request.

\subsection*{Ethics Approval and Consent to Participate Statement}

This study involved secondary analysis of de-identified, publicly available data. All methods were performed in accordance with relevant guidelines and regulations. As no new data collection from human participants was conducted by the authors, ethics approval and informed consent were not required. Ethical approval and participant consent were obtained by the original data providers, as detailed in the original publications.

\clearpage
\bibliographystyle{unsrt}
\bibliography{main}


\clearpage
\renewcommand\thefigure{S\arabic{figure}}    
\renewcommand\thetable{S\arabic{table}}    
\renewcommand\thealgorithm{S\arabic{algorithm}}    
\setcounter{figure}{0}  
\setcounter{table}{0}  
\setcounter{algorithm}{0}  

\appendix


\section*{Supplementary Materials}\label{secA0}

We compare the semantic similarity of the working alliance inventories with the transcripts using Algorithm \ref{alg:waa}. The dialogue between the patient and therapist is transcribed into pairs of turns denoted as $S^p_i$ for the patient's response turn, followed by the therapist's response turn $S^t_i$. The inventories of working alliance questionnaires are also provided in pairs: $I^p$ for the patient and $I^t$ for the therapist, each comprising 36 statements. We use sentence or paragraph embeddings to encode both the dialogue turns and the inventories, and then compute the cosine similarity between the embedding vectors of each turn and its corresponding inventory vectors. This approach yields a 36-dimensional working alliance score for each turn.

\begin{algorithm}[!h]
 \caption{Working Alliance Analysis (WAA)}
 \label{alg:waa}
 \begin{algorithmic}[1]
 \State {\bfseries }\textbf{for} i = 1,2,$\cdots$, T \textbf{do} 
\State {\bfseries } \quad Automatically transcribe dialogue turn pairs  $(S^p_i,S^t_i)$
\State {\bfseries }\quad \textbf{for} $(I^p_j, I^t_j) \in$ inventories $(I^p, I^t)$ \textbf{do} 
\State {\bfseries }\quad \quad Score $W^{p_i}_{j}$ = similarity($Emb({I^p_j}), Emb(S^p_i)$) 
\State {\bfseries }\quad \quad Score $W^{t_i}_{j}$ = similarity($Emb({I^t_j}), Emb(S^t_i)$) 
\State {\bfseries } \quad \textbf{end for}
\State {\bfseries } \textbf{end for}
 \end{algorithmic}
\end{algorithm}

For classification tasks, we concatenate the 36-dimensional working alliance scores estimated from the current turn with the sentence embedding of the turn. This combined feature vector is then fed into the Working Alliance Transformer (WAT) and Working Alliance LSTM (WA-LSTM) Algorithm \ref{alg:wat}, which are based on the Transformer architecture \cite{vaswani2017attention} and the LSTM network model \cite{hochreiter1997long} respectively. The WAT and WA-LSTM serve as sequence classifiers, taking in the sequence of feature vectors and predicting the clinical condition associated with the sequence. 

\begin{algorithm}[!h]
 \caption{Working Alliance Transformer (WAT) and LSTM (WA-LSTM)}
 \label{alg:wat}
 \begin{algorithmic}[1]
 \State {\bfseries } \textbf{Input}: a session with $T$ turns
 \State {\bfseries } \textbf{Output}: a label for psychiatric condition
 \State {\bfseries }\textbf{for} i = 1,2,$\cdots$, T \textbf{do}
\State {\bfseries } \quad Automatically transcribe dialogue turn pairs  $(S^p_i,S^t_i)$
\State {\bfseries }\quad \textbf{for} $(I^p_j, I^t_j) \in$ inventories $(I^p, I^t)$ \textbf{do} 
\State {\bfseries }\quad \quad Score $W^{p_i}_{j}$ = similarity($Emb({I^p_j}), Emb(S^p_i)$) 
\State {\bfseries }\quad \quad Score $W^{t_i}_{j}$ = similarity($Emb({I^t_j}), Emb(S^t_i)$) 
\State {\bfseries } \quad \textbf{end for}
\State {\bfseries }\quad Patient feature $x_c = concat(Emb(S^t_i), W^{t_i})$ 
\State {\bfseries }\quad Therapist feature $x_t = concat(Emb(S^t_i), W^{t_i})$ 
\State {\bfseries }\quad  Full feature $x = concat(x_t, x_c)$ 
\State {\bfseries }\quad Aggregated feature $X.append(x)$
\State {\bfseries } \textbf{end for}
\State {\bfseries } obtain prediction $y=Transformer(X)$ or $y=LSTM(X)$ 
 \end{algorithmic}
\end{algorithm}

Several neural topic models are evaluated in this study. The Neural Variational Document Model (NVDM) \cite{miao2016neural} is an unsupervised text modeling approach based on variational autoencoders. We use the Gaussian softmax construction (NVDM-GSM) variant, which achieves low perplexity and is recommended for topic modeling \cite{miao2017discovering}. Another approach, the Wasserstein-based Topic Model (WTM), utilizes Wasserstein autoencoders (WAE) to enforce a Dirichlet prior on the latent document-topic vectors \cite{nan2019topic}. We evaluate two variants of WTM: WTM-MMD, which minimizes Maximum Mean Discrepancy (MMD) for distribution matching, and WTM-GMM, which applies a Gaussian Mixture prior with Gaussian softmax. Additionally, we employ the Embedded Topic Model (ETM) \cite{dieng2020topic}, which models each word with a categorical probability distribution based on the inner product between a word embedding and a topic embedding. Finally, we utilize the Bidirectional Adversarial Training Model (BATM), which applies bidirectional adversarial training to construct a two-way projection between the document-word distribution and the document-topic distribution \cite{wang2020neural}.

The pipeline for temporal topic modeling analysis (TMM) is outlined in Algorithm \ref{alg:ttm}. For each turn in the transcript, we calculate a topic score vector with each dimension representing the likelihood of the turn belonging to a specific topic. To characterize the directional property of each turn with a particular topic, we compute the cosine similarity between the embedded topic vector and the embedded turn vector. This approach provides a measure of the topic relevance to each turn in the sequence.

\begin{algorithm}[!h]
 \caption{Temporal Topic Modeling (TTM)}
 \label{alg:ttm}
 \begin{algorithmic}[1]
 \State {\bfseries } Learned topics $T$ as references
 \State {\bfseries }\textbf{for} i = 1,2,$\cdots$, N \textbf{do}
\State {\bfseries } \quad Automatically transcribe dialogue turn pairs  $(S^p_i,S^t_i)$
\State {\bfseries }\quad \textbf{for} $T_j \in$ topics $T$ \textbf{do} 
\State {\bfseries }\quad \quad Topic score $W^{p_i}_{j}$ = similarity($Emb({T_j}), Emb(S^p_i)$) 
\State {\bfseries }\quad \quad Topic score $W^{t_i}_{j}$ = similarity($Emb({T_j}), Emb(S^t_i)$) 
\State {\bfseries } \quad \textbf{end for}
\State {\bfseries } \textbf{end for}
 \end{algorithmic}
\end{algorithm}


\section*{Supplementary Tables}\label{secA1}

\begin{table}[!h]
\caption{{Working Alliance Inventory (WAI). Shown here is the 36-item WAI client version, but the therapist version is similar.}}\label{tab:wai}
      \centering
      \resizebox{\linewidth}{!}{
\begin{tabular}{llc}
\hline
\textbf{Scale} & \textbf{Inventory item} & \textbf{Directionality} \\
\hline
BOND & I felt uncomfortable with \_. & - \\
TASK & \_ and I agreed about the things I will need to do in therapy to help improve my situation. & + \\ 
GOAL & I was worried about the outcome of the sessions. & - \\
TASK & What I was doing in therapy gave me new ways of looking at my problem. & + \\
BOND & \_ and I understood each other. & + \\
GOAL & \_ perceived accurately what my goals were. & + \\
TASK & I find what I was doing in therapy confusing. & - \\
BOND & I believe \_ liked me. & + \\
GOAL & I wish \_ and I could have clarified the purpose of our sessions. & - \\
GOAL & I disagreed with \_ about what I ought to get out of therapy. & - \\
TASK & I believe the time \_ and I were spending together was not spent efficiently. & - \\
GOAL & \_ did not understand what I was trying to accomplish in therapy. & - \\
TASK & I was clear on what my responsibilities were in therapy. & + \\
GOAL & The goals of the sessions were important for me. & + \\
TASK & I find what \_ and I were doing in therapy was unrelated to my concerns. & - \\
TASK & I feel that the things I did in therapy helped me to accomplish the changes that I wanted. & + \\
BOND & I believe \_ was genuinely concerned for my welfare. & + \\
TASK & I was clear as to what \_ wanted me to do in those sessions. & + \\
BOND & \_ and I respected each other. & + \\
BOND & I feel that \_ was not totally honest about his/her feelings toward me. & - \\
BOND & I was confident in \_'s ability to help me. & + \\
GOAL & \_ and I were working towards mutually agreed upon goals. & + \\
BOND & I feel that \_ appreciated me. & + \\
TASK & We agreed on what was important for me to work on. & + \\
GOAL & As a result of the therapy I became clearer as to how I might be able to change. & + \\
BOND & \_ and I trusted one another. & + \\
GOAL & \_ and I had different ideas on what my problems were. & - \\
BOND & My relationship with \_ was very important to me. & + \\
BOND & I had the feeling that if I said or did the wrong things, \_ would stop working with me. & - \\
GOAL & \_ and I collaborated on setting goals for my therapy. & + \\
TASK & I was frustrated by the things I was doing in therapy. & - \\
GOAL & We had a good understanding of the kind of changes that would be good for me. & + \\
TASK & The things that \_ was asking me to do did not make sense. & - \\
GOAL & I did not know what to expect as the result of my therapy. & - \\
TASK & I believe the way we were working with my problem was correct. & + \\
BOND & I feel \_ cared about me even when I did things that he/she did not approve of. & + \\
\hline
\end{tabular}
}
\end{table}


\begin{table*}[!h]
      \caption{Topic evaluations of the neural topic models (following \cite{mimno2011optimizing})
      }
      \label{tab:topic_eval} 
      \centering
      \resizebox{\linewidth}{!}{
 \begin{tabular}{ l  c  c  c  c  c  c c c }
  \toprule
 &\multicolumn{2}{c}{Anxiety} & &\multicolumn{2}{c}{Depression} & &\multicolumn{2}{c}{Schizophrenia}  \\ \cmidrule{2-3} \cmidrule{5-6} \cmidrule{8-9}
  & Topic diversity & Topic coherence &
  & Topic diversity & Topic coherence &
  & Topic diversity & Topic coherence   \\ \midrule
NVDM-GSM & 0.653 & \textbf{-380.933} & &
0.487 & -316.439 & & 
0.527 & -431.393 \\
WTM-MMD & \textbf{0.927} & -453.929 & &
0.907 & -359.964 & &
0.447 & -403.694 \\
WTM-GMM & 0.907 & -425.515 & &
0.340 & \textbf{-236.815} & &
0.467 & \textbf{-204.930} \\
ETM & 0.893 & -449.000 & &
\textbf{0.933} & -367.069 & &
\textbf{0.973} & -310.211 \\
BATM & 0.720 & -441.049 & & 
0.773 & -443.394 & &
0.500 & -337.825 \\ \hline
\end{tabular}
 }
\end{table*}


\begin{table}[!h]
      \caption{Coherence embedding evaluations of the neural topic models (following \cite{roder2015exploring})
      }
      \label{tab:emb_eval} 
      \centering
      \resizebox{\linewidth}{!}{
 \begin{tabular}{ l  c  c  c  c  c  c  c  c  c  c  c  c c c }
 \toprule
 &\multicolumn{4}{c}{Anxiety} &  &\multicolumn{4}{c}{Depression} &  &\multicolumn{4}{c}{Schizophrenia} \\ \cmidrule{2-5} \cmidrule{7-10} \cmidrule{12-15}
  & $c_v$ & $c_{w2v}$ & $c_{uci}$ & $c_{npmi}$ &
  & $c_v$ & $c_{w2v}$ & $c_{uci}$ & $c_{npmi}$ &
  & $c_v$ & $c_{w2v}$ & $c_{uci}$ & $c_{npmi}$  \\ \midrule
NVDM-GSM & 0.410 & 0.484 & -0.844 & -0.019 & & 
0.495 & 0.531 & -3.522 & -0.109 & &
0.642 & - & -1.954 & -0.065 \\
WTM-MMD & 0.340 & 0.428 & -2.827 & -0.099 & &
0.290 & 0.462 & -3.797 & -0.124 & &
0.576 & 0.751 & -0.997 & -0.036 \\
WTM-GMM & 0.353 & 0.413 & -3.259 & -0.116 & &
0.678 & 0.535 & -0.126 & -0.006  & &
0.572 & 0.774 & -1.587 & -0.050 \\
ETM & 0.413 & - & -2.903 & -0.093 & &
0.403 & - & -2.399 & -0.05 & &
0.379 & 0.864 & -7.232 & -0.199 \\
BATM & 0.352 & 0.387 & -5.056 & -0.190 & &
0.404 & 0.423 & -4.238 & -0.160 & &
0.507 & 0.816 & -9.655 & -0.343 \\ \hline
\end{tabular}
 }
\end{table}


\begin{table}[!h]
\centering
\caption{Pairwise comparison of TASK scale scores (Therapist) with statistical t-test
}
\label{stats_wa_task_t}\begin{tabular}{|c|c|c|c|c|}
\hline
 & Anxiety & Depression & Schizophrenia & Suicidality \\ \hline
Anxiety & - & & & \\ \hline
Depression & **** (1.550e-26) & - & & \\ \hline
Schizophrenia & *** (3.835e-05) & ns (1.267e-01) & - & \\ \hline
Suicidality & ns (6.521e-01) & * (1.900e-02) & ns (6.568e-02) & - \\ \hline
\end{tabular}
\end{table}


\begin{table}[!h]
\centering
\caption{Pairwise comparison of TASK scale scores (Patient) with statistical t-test
}
\label{stats_wa_task_p}
\begin{tabular}{|c|c|c|c|c|}
\hline
 & Anxiety & Depression & Schizophrenia & Suicidality \\ \hline
Anxiety & - & & & \\ \hline
Depression & **** (1.796e-25) & - & & \\ \hline
Schizophrenia & ** (1.937e-03) & **** (7.263e-17) & - & \\ \hline
Suicidality & **** (8.197e-20) & **** (2.108e-28) & **** (5.465e-14) & - \\ \hline
\end{tabular}
\end{table}


\begin{table}[!h]
\centering
\caption{Pairwise comparison of BOND scale scores (Therapist) with statistical t-test
}
\label{stats_wa_bond_t}\begin{tabular}{|c|c|c|c|c|}
\hline
 & Anxiety & Depression & Schizophrenia & Suicidality \\ \hline
Anxiety & - & & & \\ \hline
Depression & **** (1.600e-22) & - & & \\ \hline
Schizophrenia & *** (2.998e-05) & ns (3.249e-01) & - & \\ \hline
Suicidality & ns (6.004e-01) & ns (2.176e-01) & ns (3.908e-01) & - \\ \hline
\end{tabular}
\end{table}


\begin{table}[!h]
\centering
\caption{Pairwise comparison of BOND scale scores (Patient) with statistical t-test
}
\label{stats_wa_bond_p}\begin{tabular}{|c|c|c|c|c|}
\hline
 & Anxiety & Depression & Schizophrenia & Suicidality \\ \hline
Anxiety & - & & & \\ \hline
Depression & **** (2.207e-13) & - & & \\ \hline
Schizophrenia & **** (1.565e-11) & ** (3.883e-03) & - & \\ \hline
Suicidality & **** (3.627e-23) & **** (2.555e-29) & **** (3.629e-34) & - \\ \hline
\end{tabular}
\end{table}


\begin{table}[!h]
\centering
\caption{Pairwise comparison of GOAL scale scores (Therapist) with statistical t-test
}
\label{stats_wa_goal_t}\begin{tabular}{|c|c|c|c|c|}
\hline
 & Anxiety & Depression & Schizophrenia & Suicidality \\ \hline
Anxiety & - & & & \\ \hline
Depression & **** (7.186e-08) & - & & \\ \hline
Schizophrenia & * (3.010e-02) & ns (4.959e-01) & - & \\ \hline
Suicidality & **** (1.115e-10) & **** (3.532e-13) & **** (3.249e-12) & - \\ \hline
\end{tabular}
\end{table}


\begin{table}[!h]
\centering
\caption{Pairwise comparison of GOAL scale scores (Patient) with statistical t-test
}
\label{stats_wa_goal_p}\begin{tabular}{|c|c|c|c|c|}
\hline
 & Anxiety & Depression & Schizophrenia & Suicidality \\ \hline
Anxiety & - & & & \\ \hline
Depression & ns (1.177e-01) & - & & \\ \hline
Schizophrenia & *** (8.006e-04) & **** (4.519e-05) & - & \\ \hline
Suicidality & ns (2.148e-01) & ns (1.289e-01) & ns (9.510e-01) & - \\ \hline
\end{tabular}
\end{table}

\end{document}